\documentclass{article}

\usepackage{microtype}
\usepackage{graphicx}
\usepackage{subcaption}
\usepackage{booktabs} 
\usepackage{enumitem}
\usepackage{hyperref}

\usepackage[accepted]{template/icml2026}

\makeatletter
\renewcommand{\ICML@appearing}{%
  \textit{Proceedings of the ICML 2026 Workshop on Structured Probabilistic Inference
  \& Generative Modeling (SPIGM)}, Seoul, South Korea. 2026.
  Copyright 2026 by the author(s).}
\makeatother

\usepackage{amsmath}
\usepackage{amssymb}
\usepackage{mathtools}
\usepackage{amsthm}
\usepackage{xcolor}
\usepackage{tikz}

\definecolor{matblue}{HTML}{1F77B4}
\definecolor{matorange}{HTML}{FF7F0E}
\definecolor{matgreen}{HTML}{2CA02C}
\definecolor{matred}{HTML}{D62728}
\definecolor{matgray}{HTML}{7F7F7F}

\usepackage[capitalize]{cleveref}

\theoremstyle{plain}

\theoremstyle{definition}

\theoremstyle{remark}

\usepackage[textsize=tiny]{todonotes}

\icmltitlerunning{How to Spend Your  Oracle Budget: Practical Guidance for Protein Structure Prediction Models}

\begin{document}

\twocolumn[
  \icmltitle{How to Spend Your Oracle Budget: \\ Practical Guidance for Protein Structure Prediction Models}



  \icmlsetsymbol{equal}{*}

  \begin{icmlauthorlist}  
    \icmlauthor{Aleksandra Kalisz}{xxx,yyy}
    \icmlauthor{Jack Simons}{xxx}
    \icmlauthor{Krisztina Sinkovics}{xxx}
    \icmlauthor{Noam Ghenassia}{xxx}
    \icmlauthor{Shikha Surana}{xxx} \\
    \icmlauthor{Henry Moss}{zzz}
    \icmlauthor{Paul Duckworth}{xxx}
  \end{icmlauthorlist}

  \icmlaffiliation{xxx}{InstaDeep Ltd}
  \icmlaffiliation{yyy}{University of Oxford}
  \icmlaffiliation{zzz}{Lancaster University}

  \icmlcorrespondingauthor{Paul Duckworth}{p.duckworth@instadeep.com}
  
  \icmlkeywords{Generative Models, Latent Space Optimisation, Structure Prediction, Proteins, AI4Science}

  \vskip 0.3in
]



\printAffiliationsAndNotice{}  

\begin{abstract}

Foundation models for protein structure prediction remain unreliable on certain targets. External oracles can flag and correct these failures, but biological oracles are expensive, making oracle budget a critical constraint. Existing guidance methods, such as FK-steering, DPO, and Best K-of-N sampling, differ in how they spend this budget, yet no systematic comparison exists to guide method selection. To bridge this gap, we benchmark these methods alongside the recently proposed Optimisation Over Outputs (O3), which applies off-the-shelf optimisers within a generative model's latent subspace. 
We extend the usage of O3 to protein structure prediction models. Overall, our work provides the first practical reference for oracle budget-aware guidance. Our evaluation on two protein targets, calmodulin (1CLL) and E. coli aspartate transcarbamoylase (9EEH), reveals that no single method consistently dominates across all budgets and oracles. Specifically, O3 proves most effective at low oracle budgets, while FK-steering and DPO demonstrate improved performance as the budget increases. 
We distil these findings into actionable recommendations for practitioners operating under real-world oracle-budget constraints.
\end{abstract}

\section{Introduction}
\label{sec:intro}
Foundation models for structure prediction such as AlphaFold3 \cite{abramson2024accurate}, Chai-1 \cite{chai2024decoding}, and Boltz-2 \cite{passaro2025boltz} can accurately predict structures from sequence information alone. Downstream applications, including drug discovery and antibody therapeutic design, rely on these predictions being correct. In practice these predictions are not always reliable: a model may collapse onto a single conformation when several are functionally relevant \cite{waymentsteele2023predicting}, produce physically implausible geometries \cite{buttenschoen2024posebusters}, or assemble large biomolecular complexes incorrectly \cite{yin2023evaluation}. A typical procedure to flag unreliable predictions is via the usage of external oracles, which assign a score to a given prediction, e.g. a molecular dynamics simulation of conformational stability or binding energy \cite{hollingsworth2018molecular}. A natural research direction is whether such oracles can be leveraged at training or inference time to return higher scoring structures, a process referred to as \emph{guidance}. There are two dominant classes of guidance approaches, inference-time steering and fine-tuning; however their relative effectiveness is poorly understood, a gap in the literature we seek to fill.


\begin{figure*}[t]
\centering
\includegraphics[width=0.8\linewidth]{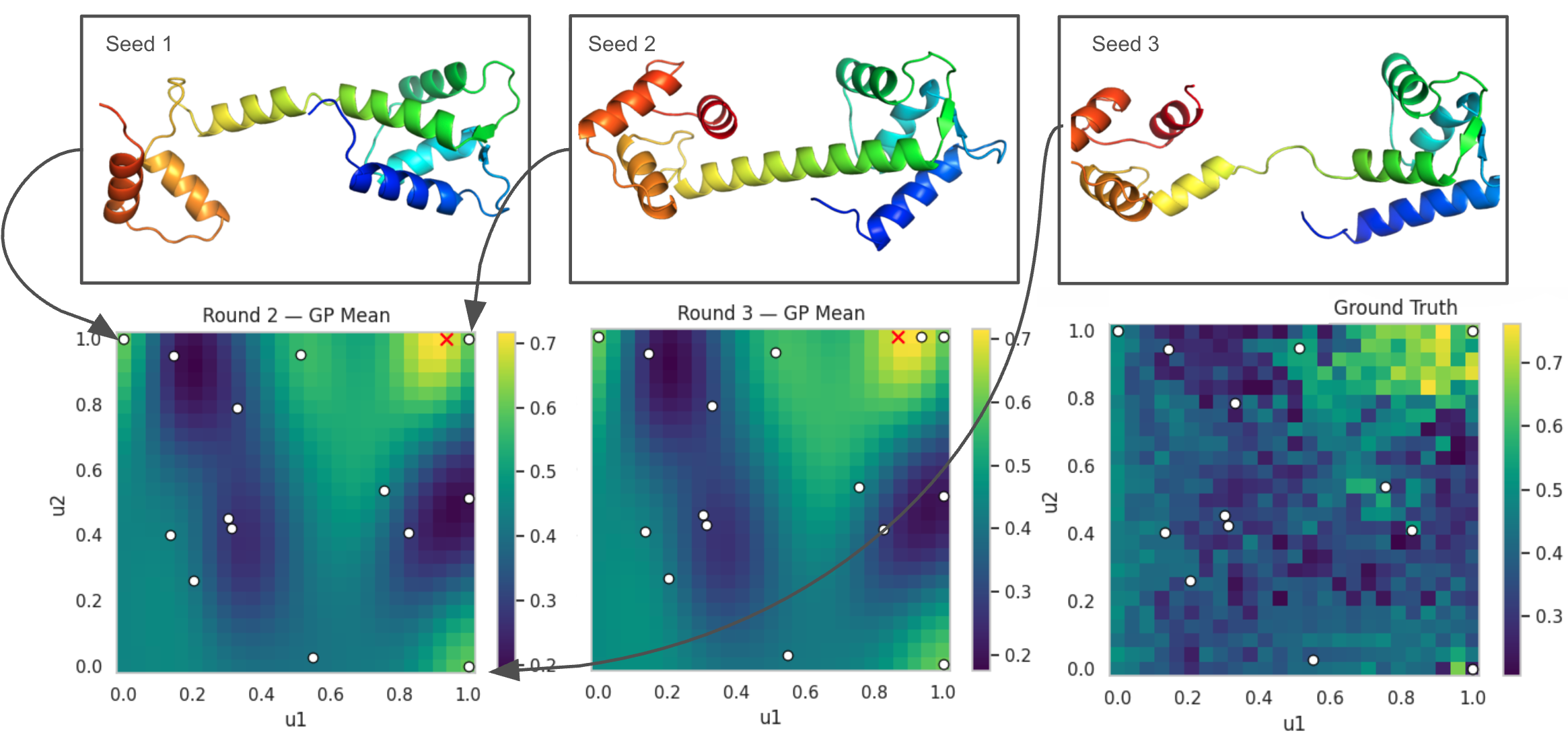}
\caption{
\textbf{Bayesian Optimisation via O3 in a Boltz-2.} $d=3$ surrogate subspace created from three Calmodulin seed samples (top row), showing the 2D $\mathcal{U}$ space (bottom row), the GP training data (white dots) and mean predictive posterior over $\mathcal{U}$, the max acquisition point (red cross) for two acquisition rounds, and a 25x25 exhaustive ground-truth grid directly evaluating the oracle (bottom right).}
\label{fig:o3_bayes_opt}
\end{figure*}

Inference-time methods, such as Feynman-Kac steering (FK-steering) \cite{singhal2025general}, sample multiple interacting particles through the generative process and resample at intermediate steps using oracle scores. Fine-tuning methods, such as Direct Preference Optimisation (DPO) \cite{rafailov2023direct}, instead update model parameters using oracle-scored samples. An even simpler approach to guidance with an oracle is Best K-of-N sampling, whereby we draw $N$ samples and return the $K$ with the highest oracle scores. All three approaches improve outputs given enough oracle evaluations, but biological oracles are usually expensive: a single molecular dynamics simulation can take days of GPU compute \cite{hollingsworth2018molecular}, and wet-lab assays are slower still.

As well as benchmarking the methods listed above, we
also provide the first application of the newly proposed Optimisation Over Outputs (O3) framework of \citet{willis2025defining} to a protein structure prediction model. O3 uses a small set of example model generations to build a low-dimensional subspace of the latent space over which standard optimisers can be directly applied. More details of the precise methodology are given in \citet{willis2025defining}.

We analyse the relative performances of O3, FK-steering, DPO, and Best K-of-N sampling applied to Boltz-2~\cite{passaro2025boltz} across a range of budgets, with the goal to improve the quality of the predicted protein structures. We find that our new application of O3 to Boltz-2 dominates at low-to-mid budgets that we evaluate on, while FK-steering and DPO perform better with larger budgets. 
This qualitative trade-off is demonstrated in our experimental study reported in \cref{sec:results}. The primary goal of this paper is straightforward: \textit{provide the first practical reference for which guidance strategies are effective under different oracle evaluation budgets.}


\section{Related Work}
\label{sec:related}
\paragraph{Protein structure prediction.}
Recent foundation models for protein structure prediction, including AlphaFold3 \cite{abramson2024accurate}, Chai-1 \cite{chai2024decoding}, and Boltz-2 \cite{passaro2025boltz}, share a common architecture: an attention-based trunk, typically a Pairformer, that processes the input sequence into single and pair representations, which is then used as conditioning information to a diffusion block over atomic coordinates. 
Throughout our experiments we use the open-source model Boltz-2. 

\paragraph{Latent space optimisation.}
Black-box optimisation describes optimisation of an objective function which is explicitly unknown \textit{a priori} but can be queried through function evaluations. These evaluations can be expensive, noisy, and derivative-free. A popular approach to such problems is Bayesian Optimisation (BO) (see e.g. \citet{garnett2023bayesian}). However, standard BO approaches begin to fail in high-dimensional or highly structured spaces. To tackle this shortcoming, several Latent space Bayesian Optimisation (LSBO) approaches have been proposed which instead perform BO in the latent space associated with a generative model. 
Examples of latent space Bayesian optimisation include VAE-BO~\cite{gomezbombarelli2018automatic}, weighted retraining~\cite{tripp2020sampleefficient}, LOL-BO~\cite{maus2022local}, and COWBOYS~\cite{moss2025return}.
Generative models often have latent spaces that are too large to apply these methods directly, fortunately, recent work from \citet{bodin2024linear} and \citet{willis2025defining} proposes an alternative approach that extracts low-dimensional subspaces from generative models, over which standard optimisers can be applied.


\vspace{-0.2cm}
\paragraph{Inference-time guidance.}
Inference-time guidance biases the sampling trajectory of a diffusion model toward a target distribution without modifying its weights. Classifier guidance \cite{dhariwal2021diffusion} perturbs the sampling trajectory by adding the gradient of an externally-trained noise-aware classifier to the denoising step, while classifier-free guidance \cite{ho2022classifierfree} trains a single model on both conditional and unconditional inputs and combines their score estimates at inference time. 
We choose Feynman-Kac steering \cite{singhal2025general} as a baseline for this work since it does not require propagating gradients from the oracle, and does not require an auxiliary network. 

\vspace{-0.2cm}
\paragraph{Reward-based fine-tuning.}
DDPO \cite{black2023training} casts sampling as a multi-step Markov decision process and applies policy-gradient updates with the oracle as reward; DRaFT \cite{clark2023directly} backpropagates the oracle gradient through the sampling chain when the oracle is differentiable. DPO \cite{rafailov2023direct} sidesteps explicit reward modelling by fine-tuning on pairs of oracle-scored samples to increase the relative likelihood of the preferred one, and Diffusion-DPO \cite{wallace2023diffusion} adapts this objective to the diffusion likelihood. We choose DPO as our fine-tuning baseline because it requires only sample pairs and oracle scores. 

Typically, biological oracles are non-differentiable, meaning that gradient information is not available, making many common guidance approaches unsuitable.

\section{Guidance Methods}
\label{sec:method}
We denote our frozen generative model $g: \mathcal{Z} \to \mathcal{X}$, that maps random noise  $z\sim p_0$ from within a latent space $z\in\mathcal{Z}$ to structures $x \sim p_1$ in data space $x\in\mathcal{X}$. An oracle $r: \mathcal{X} \to \mathbb{R}$ scores each generated output. In our experiments $g$ is the protein structure prediction model Boltz-2~\cite{passaro2025boltz} with $\mathcal{Z} = \mathcal{X} = \mathbb{R}^{3n}$ for an $n$-atom system and $p_0 = \mathcal{N}(0, I)$. The goal of guidance is to use this model to generate a batch of $K$ candidate outputs whose scores under $r$ should be as high as possible, under the constraint that we only make $N$ oracle queries. 
We will now introduce three guidance methods, explaining the key modifications and design choices that were required to apply them to protein structure prediction models.

\subsection{Bayesian Optimisation via O3}
\label{sec:method:surrogate}
We now describe how the O3 framework of \citet{willis2025defining} is used to guide the generative process of Boltz-2. The algorithm relies on two key components, each necessitating a number of oracle queries: (i) the construction of a suitable low-dimensional subspace $\mathcal{U}$ within which generation is constrained (the first $M$ queries), and (ii) a resource-efficient optimization procedure to effectively explore this subspace (the remaining $N-M$ queries).

\textbf{(i) Constructing $\mathcal{U}$}. Given a total budget of $N$, we first sample $M < N$ initial structures from $g$ and score them with $r$. Exploiting the determinism of $g$, we collect the $d$ latents corresponding to the $d$ highest scoring structures\footnote{O3 requires a deterministic generative model, and so we converted the stochastic Boltz-2 generation process to its equivalent probability-flow formulation ~\cite{karras2022elucidating} and disable stochastic SE(3) augmentation.} as \emph{seeds}, $Z = [z_1, \dots, z_d]^\top$. Here $d$ determines the dimension of the resulting optimisation space, $d-1$. We then follow the guidelines of O3~\cite{willis2025defining}, and construct $\mathcal{U}\subset[0,1]^{d-1}$ via a \emph{surrogate chart} $\phi: \mathcal{U} \to \mathcal{Z}$ that decodes each $u \in \mathcal{U}$ to a corresponding latent $z$ through two maps (see \cref{alg:decode}). The first, a Knothe--Rosenblatt transform~\cite{knothe1957contributions, rosenblatt1952remarks} $\phi: \mathcal{U} \to \mathbb{S}^{d-1}_+$, maps $u$ to a non-negative weight vector $w$ on the simplex. The second, a LOL projection~\cite{bodin2024linear} $\ell(w, Z) := w^\top Z$, projects these weight vectors back onto the support of our isotropic Gaussian prior $p_0$. 
This is effective because the original seeds themselves were drawn from $p_0$ and the weights are on the simplex. 
Once $d$ seeds are chosen, the subspace is fixed, and never re-built. Optimisation occurs within $\mathcal{U}$, where $\mathcal{U}$ has dimensionality $d-1$ and crucially is independent of the dimensionality of $\mathcal{Z}$, thus opening up opportunities for efficient optimization. 
In Appendix \cref{fig:boltz_interp} we select two seeds $d=2$, and linearly interpolate between them using 5 intermediate weights, displaying
their corresponding generations $g(\cdot)\sim p_1$. 

\begin{algorithm}[!ht]
\caption{Mapping from $\mathcal{U}$ to protein structures}
\label{alg:decode}
\begin{algorithmic}[1]
  \STATE $u = [u_1, \dots, u_{d-1}]^\top$ \hfill // selected by optimiser
  \STATE $w \gets \phi(u)$ \hfill // Knothe--Rosenblatt transform
  \STATE $z \gets \ell(w, Z) := w^\top Z$ \hfill // LOL projection
  \STATE $x \gets g(z)$ \hfill // decoded structure, ready for scoring
\end{algorithmic}
\end{algorithm}

\textbf{(ii) Bayesian optimization over $\mathcal{U}$.} Given a subspace $\mathcal{U}$, we use Bayesian optimisation~\citep[BO]{garnett2023bayesian} --- the gold-standard for low-dimensional black-box optimisation --- to efficiently utilise our remaining $N-M$ oracle budget. We begin by fitting a Gaussian process (GP) ~\citep{williams2006gaussian} surrogate model on $d+2$ initial points, where $d$ of them are the seed latents projected onto $\mathcal{U}$ and the remaining $2$ are sampled randomly from within $\mathcal{U}$. Each subsequent round uses Log Expected Improvement \citep{ament2023unexpected} to drive data acquisition, each time updating the GP model with the newly acquired point. 
In order to provide additional training data, we also project the $d$ seeds into the $\mathcal{U}$ subspace using the surrogate chart $\phi$, and include those as additional training data points (as demonstrated in \cref{fig:o3_bayes_opt}). 
We use an RBF kernel, constant mean function, single task GP, and Monte Carlo acquisition function sampler from BOTorch~\citep{balandat2020botorch}.

\subsection{FK-Steering} 
Feynman-Kac steering \cite{singhal2025general} is an inference-time method which uses the signal from an oracle to alter the sampling trajectory towards high-reward regions of the target distribution via Sequential Monte Carlo. The procedure generates $M$ interacting stochastic processes called particles. Particles are resampled proportionally to weights $w$ calculated for each particle $i$ using the oracle reward $r$ evaluated on the denoised predictions from $x_t^i$ (for denoising timestep~$t$):

\begin{equation}
\nonumber
    w(x^i_t) = \frac{g(x^i_t \mid x^i_{t_{\text{prev}}})}{g_{\text{proposal}}(x^i_t \mid x^i_{t_{\text{prev}}})} \lambda \exp(r(x^i_t) - r(x^i_{t_{\text{prev}}} ))
\end{equation}

where $\lambda$ controls the scale of the reward signal, and $g_{\text{proposal}}(x_t \mid x_{t_{\text{prev}}})$ is the proposal transition kernel. In our experiments, since most high-quality biological oracles are non-differentiable, we use the original \textsc{Boltz-2} transition kernel $g(x^i_t \mid x^i_{t_{\text{prev}}})$, demonstrating the gradient-free nature of FK-steering.
The rewards are calculated only at the resampling steps and there are several ways to combine them across the reward trajectory (difference, max, average); in this work we use the reward difference, which favours particles with increasing rewards.

To restrict to an oracle budget of $N$ and return a batch of $K$ distinct samples, we require $M\geq K$ particles, each of which has access to $N/M$ oracle calls. Practically, this ratio determines the number of resampling steps in our algorithm. Boltz-2 employs a particular noising schedule, which stops injecting noise in the last quarter of the trajectory in order to avoid atomic jitter. Since FK-steering relies on the diversity of particles, we only resample at intermediate denoising steps where Boltz-2 adds positive noise. 
We call the oracle once at the start of the denoising process, once at the end, and spread the remainder of the $N/M - 2$ budget across the first $3/4$ of the denoising trajectory.

\subsection{DPO}
Unlike our other approaches, Direct Preference Optimisation (DPO) ~\citep{rafailov2023direct} fine-tunes and directly modifies the generative model parameters based on oracle-scored sample pairs, with the aim of increasing the likelihood of generating high-scoring structures. DPO does not require a differentiable oracle (no explicit reward modeling is required, rather just pairs of structures and their relative oracle rankings) making it amenable to common biological oracles. However, we find that a large oracle budget is required to support effective fine-tuning at foundation model-scale.

\paragraph{Diffusion-DPO training objective.} DPO updates the model's weights $\theta$ in a direction that maximises the margin between the implicit rewards of preferred and rejected samples under a Kullback-Leibler regularised policy, as measured by the DPO loss:
\begin{equation*}
    \scalebox{0.88}{$
        \mathcal{L}_{\text{DPO}}(\theta) = -\mathbb{E}_{(\mathbf{x}_w, \mathbf{x}_l)}
        \left[
            \log \sigma\!\left(
                \beta \left(
                    \log \frac{p_\theta(\mathbf{x}_w)}{p_{\mathrm{ref}}(\mathbf{x}_w)}
                    - 
                    \log \frac{p_\theta(\mathbf{x}_l)}{p_{\mathrm{ref}}(\mathbf{x}_l)}
                \right)
            \right)
        \right]
    $},
\end{equation*}
where $\beta > 0$ controls the strength of the KL penalty against moving the new model $p_{\theta}$ too far from the pre-trained reference model $p_{\mathrm{ref}}$ (initialized at the pre-trained Boltz-2 checkpoint), and $\sigma$ represents the sigmoid function. Preference pairs always contain a preferred structure $\mathbf{x}_w$ and a rejected structure  $\mathbf{x}_l$, each sampled from our $N$ scored structures and lie above and below the median oracle score respectively. Obtaining log-likelihoods from diffusion models is prohibitively expensive~\citep{song2020score}, we follow \citet{wallace2023diffusion}, and approximate the log-likelihood ratio in the DPO loss using differences in the individual training losses ($\mathcal{L}_{\mathrm{DSM}}$) of the reference and fine-tuned model as $ \log (p_\theta(\mathbf{x})/p_{\mathrm{ref}}(\mathbf{x})) 
    \approx 
    \mathcal{L}^{\mathrm{ref}}_{\mathrm{DSM}}(\mathbf{x}) - \mathcal{L}^{\theta}_{\mathrm{DSM}}(\mathbf{x})$.


\paragraph{Online and offline DPO.} We evaluate two variants of DPO that differ in how oracle queries are distributed across training. The \emph{offline} variant draws all $N$ structures in a single batch, forms preference pairs once, and trains to convergence on the resulting fixed dataset. The \emph{online} variant interleaves sampling and training, at each of the $E$ epochs: we generate $N/E$ structures from the current model, score the structures with the oracle, form preference pairs, update parameters for one epoch; until a total budget of $N = (N/E) \times E$ queries has been utilised. Critically, the reference model $p_{\mathrm{ref}}$ is reset to the current policy at the start of each epoch, thus softening the KL regularisation. However, this resetting weakens the influence of the pre-trained model --- a tradeoff we investigate in  Section~\ref{sec:results:dpo}.

\section{Experimental Setup}
\label{sec:setup}
\begin{figure*}[!t]
\centering
\includegraphics[width=\linewidth]{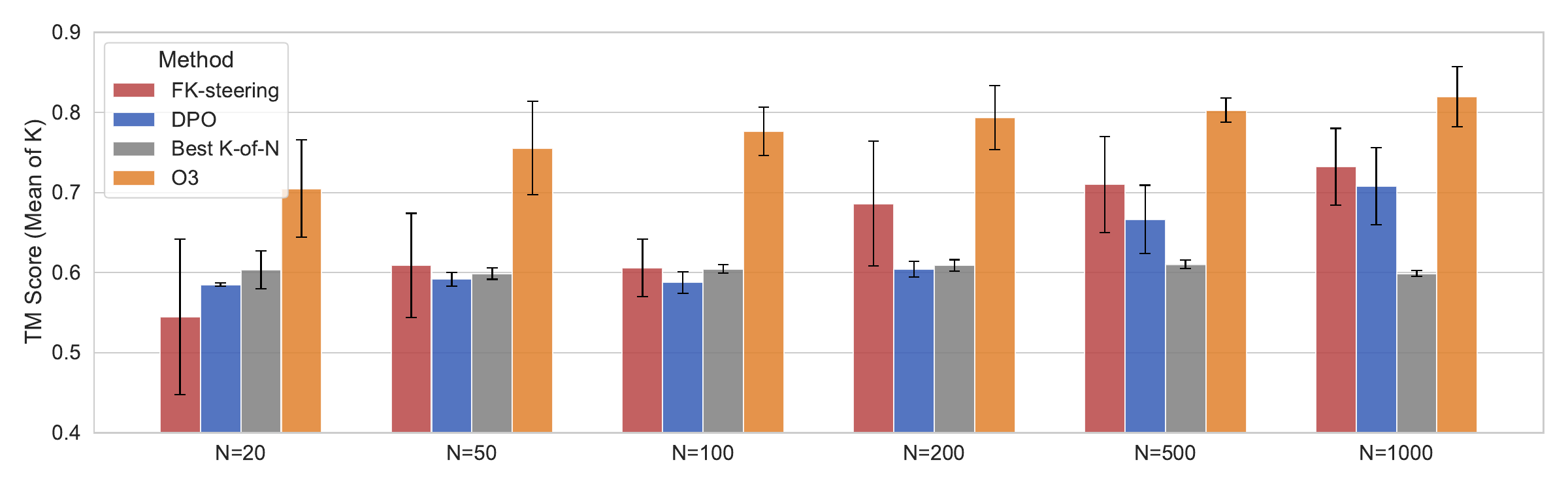}
\caption{\textbf{Mean-of-$K$ TM-score across oracle budgets.} Comparison of FK-steering, DPO (in the online setting), Best K-of-N, and O3 on 1CLL across the six $(N, K)$ configurations of \cref{sec:setup}. Bars are the mean over $5$ random seeds; error bars are $\pm 1$ std. Higher is better. O3 method dominates at budgets evaluated, while FK-steering and DPO improve steadily with $N$.}
\label{fig:max_top_k_across_n}
\end{figure*}

As our main experiment, we benchmark the four methods by comparing the predicted structures for the calmodulin protein from its amino acid sequence (of length 144). As ground truth, we compare against the 1CLL structure~\cite{chattopadhyaya1992calmodulin} consisting of 1184 atom coordinates from the Protein Data Bank (PDB)~\cite{berman2000pdb}. 
Additionally, we benchmark on E. coli aspartate transcarbamoylase (PDB: 9EEH), with results in Appendix~\ref{app:9eeh}. 
For all experiments, the base model we use, $g$, is Boltz-2~\cite{passaro2025boltz} and for 1CLL $\mathcal{Z} = \mathbb{R}^{3552}$, and 9EEH $\mathcal{Z} = \mathbb{R}^{21,696}$. 
All methods are permitted $N$ oracle calls in total and return a batch of $K$ candidate structures.
Next, we describe the TM-score oracle used in the main experiment. 

\begin{figure}[H]
\centering
\includegraphics[width=\linewidth]{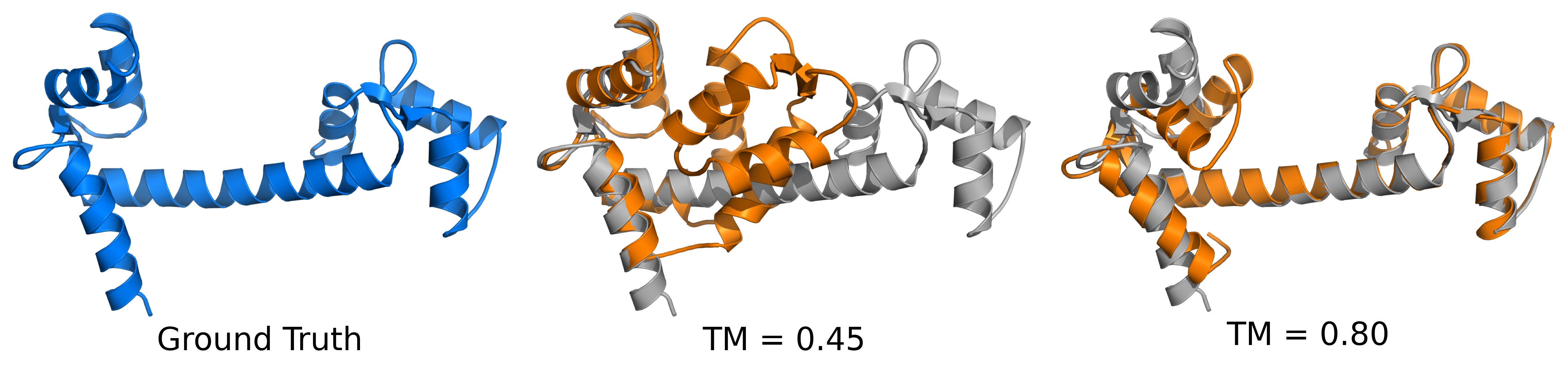}
\caption{\textbf{TM-score on 1CLL.} Left: the 1CLL crystal structure (blue). Centre: a vanilla Boltz-2 sample (orange) overlaid on the ground truth (grey) at TM-score $0.45$. Right: a guided sample at TM-score $0.80$. Structure pairs with TM-score above $0.5$ are considered to share the same fold.}
\label{fig:1cll-tm}
\end{figure}

\paragraph{Oracle.}
The reward, $r$, is the TM-score~\cite{zhang2004scoring} of a generated structure $x$ against the ground-truth 1CLL backbone $x_{\text{GT}}$. To compare $x$ with $x_{\text{GT}}$, we find the optimal rigid alignment, namely the rotation and translation of $x$ that minimise the distances to the corresponding residues of $x_{\text{GT}}$. The TM-score is then
\begin{equation*}
r(x) = \max \frac{1}{n} \sum_{i=1}^{n_{\text{ali}}} \frac{1}{1 + (d_i(x, x_{\text{GT}}) / C)^2},
\end{equation*}
where $n$ is the number of residues in the ground truth structure (144 for 1CLL); $n_{\text{ali}} \leq n$ is the number of residues from $x_{\text{GT}}$ matched to a residue in $x$ by the optimal alignment, with residues that have no nearby counterpart excluded from the sum; $d_i$ is the distance between the $i$-th aligned residue pair\footnote{it is a distance between the C$_\alpha$ atoms of the two residues.}; $C = 1.24\sqrt[3]{n - 15} - 1.8$ is a length-dependent normalisation chosen so that random structure pairs yield approximately constant TM-score; and the maximum is over the rigid alignments described above. Scores lie in $(0, 1]$, with values above $0.5$ indicating the same fold and below $0.2$ indicating random pairs; \cref{fig:1cll-tm} shows examples on 1CLL. We use TM-score because it is relatively fast to compute, thus enabling our extensive experimentation. However, we emphasise that our analysis exposes trade-offs between the studied methodologies in settings where indeed the oracle is expensive or non-differentiable, a setting common in biology.

For PDB: 9EEH experiments, we use the MolProbity oracle, described in Appendix \ref{app:9eeh:MP}.

\paragraph{Budget settings.}
We evaluate each method at six $(N, K)$ configurations spanning low to moderate oracle-call budgets: $(20, 2)$, $(50, 5)$, $(100, 10)$, $(200, 20)$, $(500, 50)$, and $(1000, 100)$. We are specifically interested in $K > 1$ settings throughout because downstream stages of biological design pipelines, such as wet-lab assay screening, often test batches of candidates rather than committing to a single proposal.

\paragraph{Evaluation metrics.}
At each $(N, K)$ configuration, we report two performance metrics: the \textit{max-of-$K$} score, the best single structure returned, and the \textit{mean-of-$K$} score, the average score across the returned batch. The former measures single-shot design quality; the latter measures batched-screening quality. Each plot in \cref{sec:results} reports one of the two, with the complementary results presented in our appendices. All experiments are repeated over $5$ random seeds unless stated otherwise.

\section{Results}
\label{sec:results}
\cref{sec:results:budgets} discusses the comparison of the four methods on PDB: 1CLL target, across six $(N, K)$ configurations introduced in \cref{sec:setup}. \cref{sec:results:o3,sec:results:fk,sec:results:dpo} elaborate on method-specific findings and ablations for O3, FK-steering, and DPO, respectively. PDB: 9EEH results are provided in Appendix \ref{app:9eeh}. 

\subsection{Model Comparison Across Budgets}
\label{sec:results:budgets}
\cref{fig:max_top_k_across_n} reports the mean-of-$K$ TM-score for our four methods (and Appendix \cref{app:1cll:all} presents the max-of-$K$ results). O3 outperforms all baselines at low-mid budgets ($N \leq 1000$) with the mean value plateauing at $\sim 0.81$. Best K-of-N is roughly flat at $\sim 0.60$ across all budgets. FK-steering and DPO improve with increased $N$, but neither are competitive with simple Best K-of-N at low budgets ($N \leq 100$), with FK-steering achieving only $\sim 0.55$ at $N=20$. From $N=200$ to $N=1000$, FK-steering performs better, reaching $\sim 0.73$ at $N=1000$; DPO improves with $N$ but does not match FK-steering or O3 within our range of settings, reaching $\sim 0.71$ at $N=1000$. O3 is the only method that meaningfully improves on the Best K-of-N baseline at low oracle budgets ($N \leq 100$), supporting the case for example-defined latent-subspace optimisation when oracle calls are scarce. Appendix \ref{app:9eeh:results} reports a model comparison across budgets on a different protein (PDB:9EEH) and MolProbity oracle (Appendix \ref{app:9eeh:MP}).

\subsection{O3}
\label{sec:results:o3}

\begin{figure}[H]
\centering
\includegraphics[width=\linewidth]{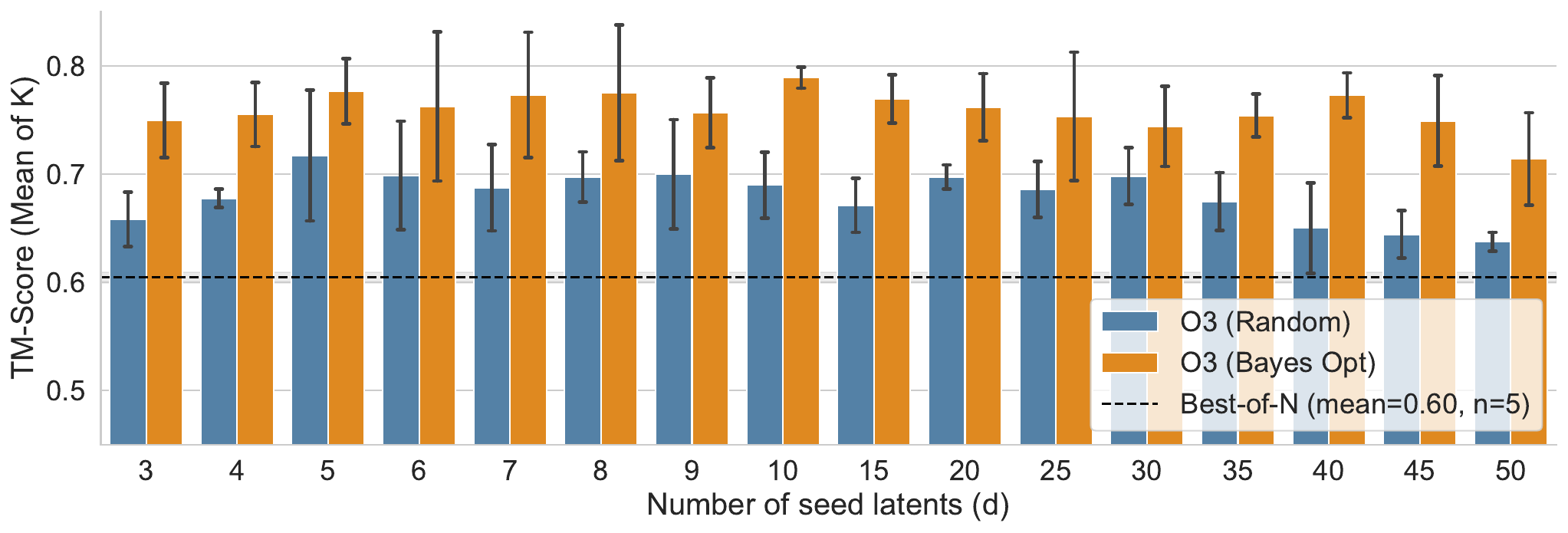}
\caption{\textbf{Bayesian optimisation versus random sampling in the O3 subspace.} Mean-of-$K$ TM-score at $N=100$ for O3 with Bayesian optimisation and O3 with uniform random sampling in the same subspace, on 1CLL. The Best K-of-N baseline is invariant under our fixed $K/N$ ratio. Bars are means over $3$ random seeds; error bars are $\pm 1$ std.}
\vspace{-0.5cm}
\label{fig:o3_random_vs_bo}
\end{figure}

\cref{fig:o3_random_vs_bo} isolates the optimisation step in O3, comparing Bayesian optimisation in the subspace against uniform random sampling in the same subspace for $N=100$. 
We create the subspace $\mathcal{U}$ by selecting to best $d$ seeds from $\mathcal{M}$. Random sampling in $\mathcal{U}$ is shown to improve over the Best K-of-N sampling baseline, demonstrating the value of the low-dimensional subspace. Furthermore, the superior performance of O3 seen in \cref{fig:max_top_k_across_n} comes from a combination of both the optimiser and the subspace construction.

\begin{figure}[h]
\centering
\includegraphics[width=\linewidth]{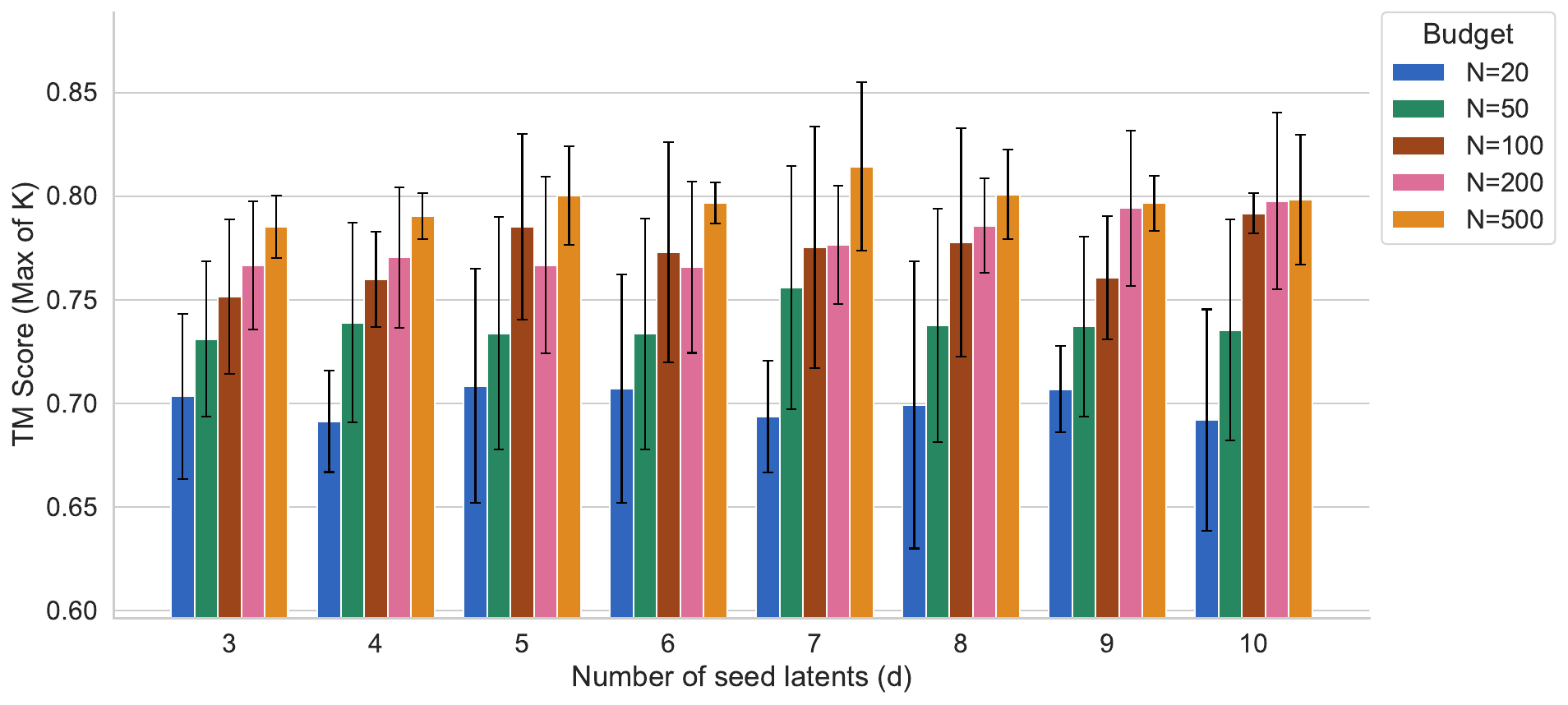}
\caption{\textbf{Performance for different O3 subspace dimensions.} Max-of-$K$ TM-score for O3 with Bayesian optimisation, across subspace dimensions $d$ for different budgets, on 1CLL. The best-performing $d$ varies with the budget. Bars are means over $3$ random seeds; error bars are $\pm 1$ std.}
\label{fig:o3_bo_latents}
\end{figure}

In the experiment summarised by \cref{fig:o3_bo_latents} we sweep across subspace dimensions $d$ for different available budgets. The key takeaway is that best-performing $d$ varies with the budget --- different oracle-call regimes favour different subspace dimensions. For example, when $N=20$ the best-performing dimension is $d=6$; however, for $N=200$ dimension $d=10$ yields better predictions.
It is clear there is a trade-off between the richness of the O3 subspace, guaranteed by increasing the number of seed latents $d$, and the resulting difficulty of the optimisation task in that subspace. Additional O3 ablations and results are presented in Appendices \ref{app:1cll:o3} and \ref{app:9eeh:results:o3}.

\subsection{FK-steering}
\label{sec:results:fk}
\begin{figure}[h]
    \centering
    \includegraphics[width=\linewidth]{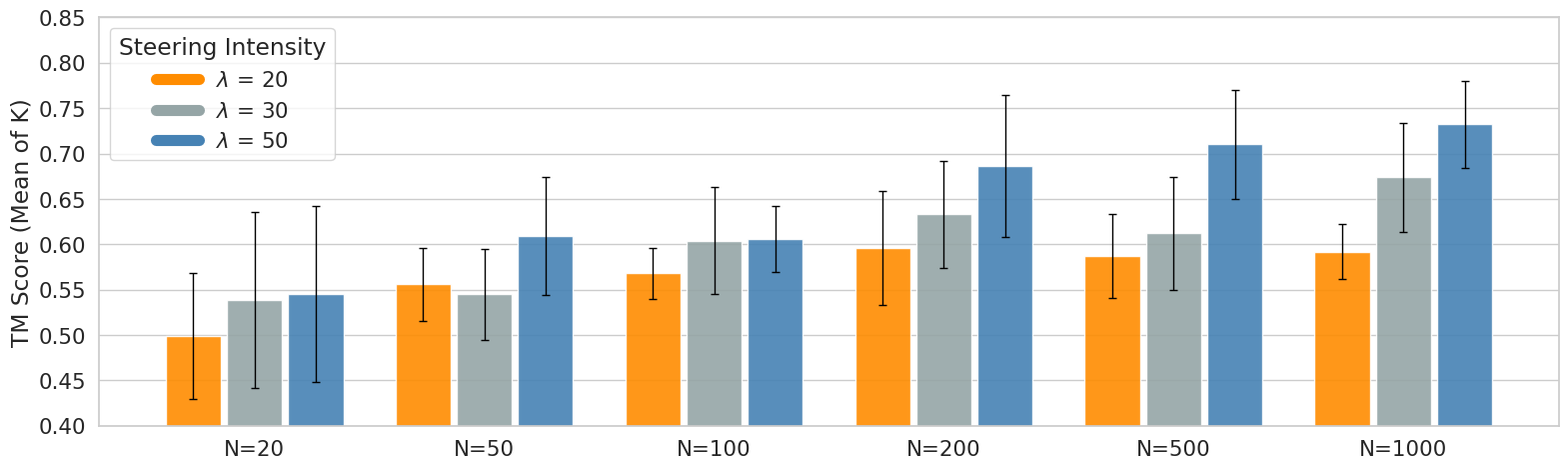}
    \caption{\textbf{Effect of FK-steering $\lambda$ hyperparameter across different budgets $N$}.  Mean-of-$K$ TM-score achieved by FK-steering on 1CLL. Bars are means over $10$ random seeds; error bars are $\pm 1$ std. Higher $\lambda$ increases the signal from the rewards and leads to better performance across budgets.}
    \label{fig:fk_lambda_mean_k}
\end{figure}

We ablate three of FK-steering's hyperparameters: the number of particles, $K$, the number of resampling steps $N/K$, and the reward-scaling coefficient $\lambda$. \cref{fig:fk_lambda_mean_k} shows that, with $K$ and $N$ fixed, increasing $\lambda$ improves performance across all budgets, however, it likely hinders output diversity.
A high $\lambda=50$ amplifies the signal from the oracle, forcing the algorithm to greedily resample the best-performing particles across the trajectory, driving the mean-of-$K$ and max-of-$K$ (see \cref{fig:fk_lambda_max_k}) closer. In \cref{fig:fk_NK_ablation} we vary $K$ per fixed budgets of $N$ and find that increasing the number of resampling steps $N/K$ at the cost of particle population $K$ generally yields better performance at higher oracle budgets. Additional FK-Steering ablations are available in Appendix~\ref{app:1cll:FK}.

\subsection{DPO}
\label{sec:results:dpo}

\begin{figure}[h]
\centering
\includegraphics[width=\linewidth]{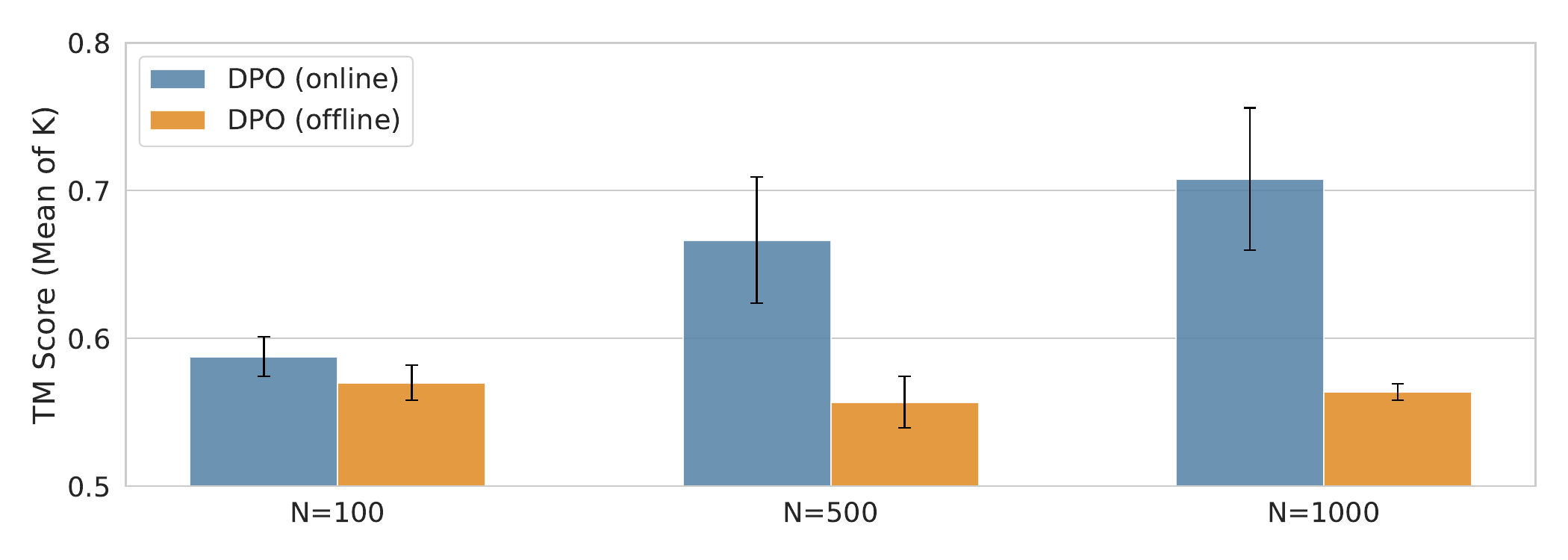}
\caption{\textbf{Online vs.\ offline DPO across oracle budgets.} Mean-of-$K$ TM-score achieved by online and offline DPO on 1CLL, for oracle budgets $N \in \{100, 500, 1000\}$. Bars are means over 5 random seeds; error bars show $\pm 1$ std. Online DPO improves consistently with $N$, while offline DPO remains flat across budgets.}
\label{fig:online-offline-dpo}
\end{figure}

\cref{fig:online-offline-dpo} compares online and offline DPO across oracle budgets. (Max-of-$K$ counterpart of this result is reported in Appendix~\ref{app:1cll:dpo}).
Online DPO improves steadily with $N$, reaching a mean TM-score of $0.708$ and a max TM-score of $0.783$ at $N=1000$. In contrast, offline DPO shows little sensitivity to budget, plateauing at $0.55$--$0.57$ across all budgets. The gap between the two variants widens with scale: negligible at $N=100$, it becomes substantial at $N=500$. These results suggest that the gains from DPO arise primarily from adaptive on-policy resampling, rather than from additional optimisation steps alone, consistent with prior observations that online preference optimisation methods can significantly outperform offline counterparts under a fixed budget ~\citep{tang2024understandingperformancegaponline}. We also find that training hyperparameters such as the preference batch size and the number of samples per epoch can meaningfully affect performance at large $N$, though a controlled study isolating each factor is left for future work.

\vspace{-1em}
\section{Conclusions}
\label{sec:conclusion}

Notably, our work represents the first application of O3 to protein structure prediction. We demonstrate that this novel approach equips practitioners with a highly effective tool, yielding exceptional results particularly in budget-constrained tasks.
Using Boltz-2 and two protein targets along with the TM-score and MolProbity oracles, we compare O3, FK-steering, DPO, and Best K-of-N sampling under oracle-budget constraints, thus providing a practical reference for budget-aware guidance in biological settings. 
Across the budgets we evaluated, O3 dominates the performance. At $\le 100$ queries, it is the only method that meaningfully improves on the Best K-of-N baseline, while FK-steering and DPO both steadily improve with $N$. As the only method that updates the model parameters, DPO is the natural candidate at substantially larger budgets than we evaluate here. 

The conclusion from this work can be summarised in the following piece of practical advice: use O3 at constrained oracle budgets. FK-steering can be competitive at moderate budgets, and reach for DPO once budgets are large enough to fine-tune. 
An additional practical consideration is batch size and required batch diversity. DPO amortises its training budget into the model weights, making sampling large subsequent batches cheap, whereas inference-time and search methods usually discard oracle values afterwards. 
Future work will include additional protein targets paired with other, potentially expensive and non-differentiable, oracles.



\newpage
\section*{Impact Statement}

Our work is a methodological comparison of guidance methods for protein-structure foundation models, aimed at helping practitioners spend limited biological oracle budgets effectively, including for downstream applications such as drug discovery and therapeutic design. As with all general-purpose tools for protein design, the same techniques could in principle be applied to harmful targets; our experiments use only the publicly deposited calmodulin structure 1CLL and E. coli aspartate transcarbamoylase structure 9EEH and pose no specific dual-use risk in themselves.

\bibliography{references}
\bibliographystyle{template/icml2026}

\newpage
\appendix
\onecolumn

\counterwithin{figure}{section}
\counterwithin{table}{section}

\section{Additional Results on PDB: 1CLL}\label{app:1cll}

\subsection{Complementary Model Comparison Across Budgets}\label{app:1cll:all}
\begin{figure*}[h]
\centering
\includegraphics[width=0.8\linewidth]{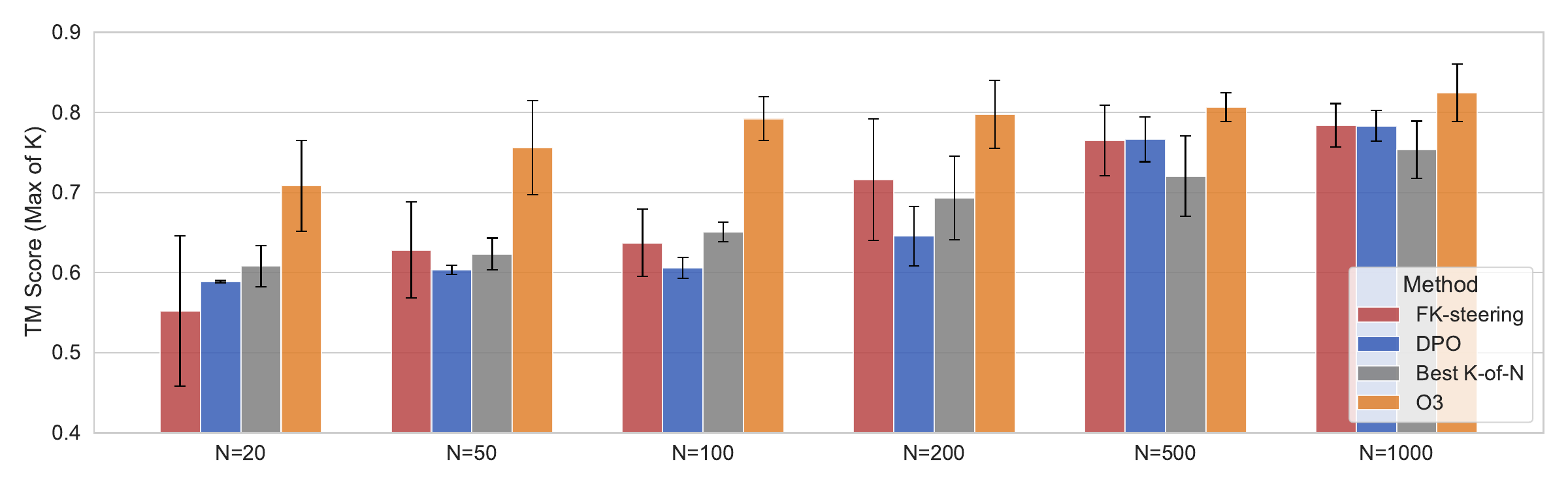}
\caption{\textbf{Max-of-$K$ TM-score across oracle budgets.} Comparison of FK-steering, DPO, Best K-of-N, and O3 on 1CLL across the six $(N, K)$ configurations of \cref{sec:setup}, reporting the maximum TM-score across the returned batch of $K$ candidate structures. Bars are the mean over $5$ random seeds; error bars are $\pm 1$ std. Higher is better.}
\label{fig:top_k_max_appendix}
\end{figure*}

\subsection{Additional O3 Details and Results on PDB: 1CLL}\label{app:1cll:o3}
\vspace{0.3cm}
\subsubsection*{O3 Experimental Details}
\vspace{-0.7cm}
\begin{table}[h]
  \centering
  \caption{%
  Experimental configurations for O3 protein structure optimisation
      (Boltz-2 generator, TM-score oracle on 1CLL).
      $N$: total Oracle budget;
      $K$: top-$K$ batch selection;
      $M$: initial samples filtered to create subspace $\mathcal{U}$;
      $d$: number of seed latents used to create the $d-1$ dimensional $\mathcal{U}$;
      $d+2$: initial points used to fit the Gaussian process in $\mathcal{U}$;
      $n_{\text{rounds}}$: number of BO acquisition rounds is set to $N-M-2$ for filtered seed-latents (and $N-(d+2)$ for random seed latents); 
      Wall-clock time of experiments run using a H100 GPU with 80Gb RAM. 
  }
  \label{tab:boltz_tm_configs}
  \begin{tabular}{lrr| rrrc}
  \toprule
  Task & $N$ & $K$ & $M$ & d & $n_{\text{rounds}}$  & wall-clock time (mins) \\
  \midrule
  \texttt{n20\_k2}     &    20 & 6 & 10 & 5 & 8   &   4.5 ($\pm$2.7)\\
  \texttt{n50\_k5}     &    50 & 7 & 25 & 7 & 23  &  6.1 ($\pm$0.4)\\
  \texttt{n100\_k10}   &   100 & 10 & 50 & 5 & 48 &  10.6 ($\pm$ 1.4) \\
  \texttt{n200\_k20}   &   200 & 10 & 100 & 10 & 98  &  20.6 ($\pm$8.6)\\
  \texttt{n500\_k50}   &   500 & 7 & 250 & 30 & 248 &  48.4588  ($\pm$2.2)\\
  \texttt{n1000\_k100} &  1000 & 100 & 400 & 50 & 598  &  153.5 ($\pm$1.6)\\
  \bottomrule
  \end{tabular}
\end{table}

\vspace{-0.2cm}
\subsubsection*{LOL Interpolation}
\begin{figure*}[h]
\centering
\includegraphics[width=0.75\linewidth]{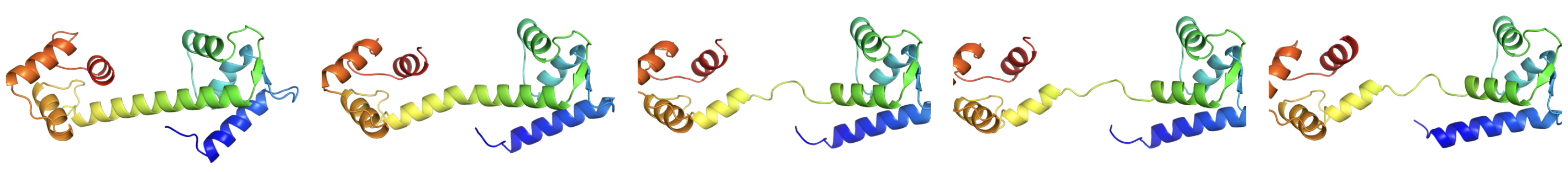}
\caption{
\textbf{LOL interpolation trajectory} in a Boltz-2 $d=2$ surrogate subspace of two Calmodulin examples, using 5 intermediate values of $(w, 1-w)$ for $w \in (0,1)$.
Structures vary smoothly along the interpolation while remaining within the support of Boltz-2. For more details see \cite{bodin2024linear}. }
\label{fig:boltz_interp}
\end{figure*}

\subsubsection*{Complementary O3 Metrics}
\begin{figure}[H]
\vspace{-0.3cm}
\centering
\begin{subfigure}{0.49\linewidth}
  \centering
  \caption{}
  \includegraphics[width=\linewidth]{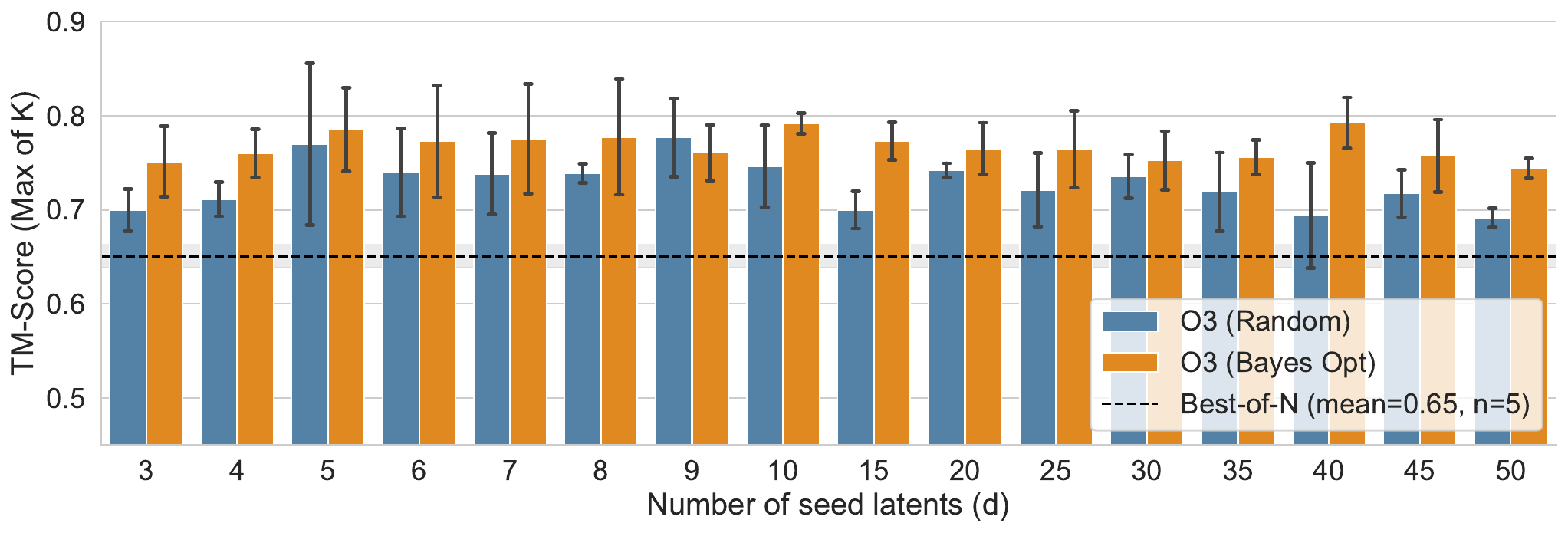}
  \label{fig:o3_random_vs_bo_max}
\end{subfigure}
\hfill
\begin{subfigure}{0.49\linewidth}
  \centering
  \caption{}
  \includegraphics[width=\linewidth]{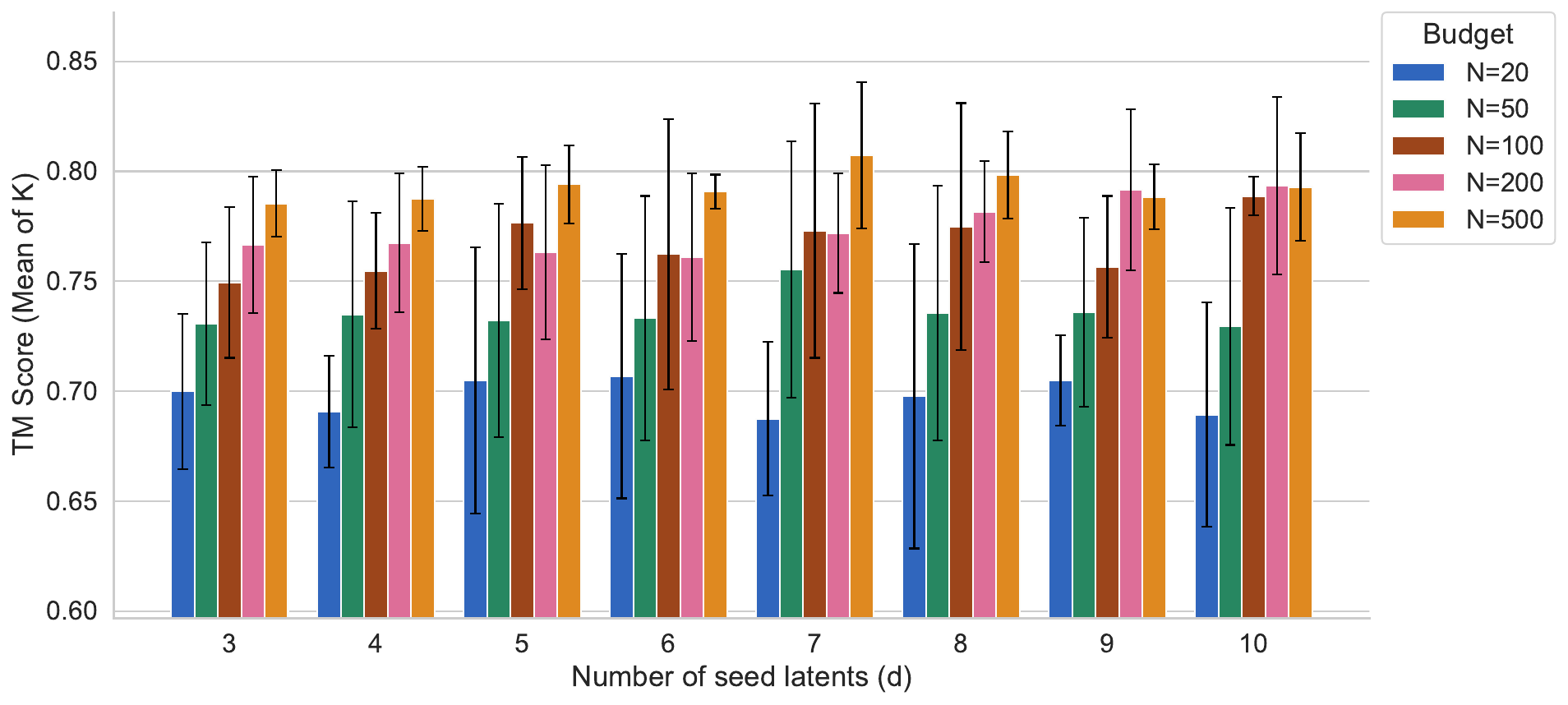}
  \label{fig:o3_bo_latents_mean}
\end{subfigure}
\vspace{-0.3cm}
\caption{\textbf{Complementary O3 metrics.} (a) Max-of-$K$ counterpart to \cref{fig:o3_random_vs_bo}: BO versus random sampling at $N=100$ and $K=10$. \newline (b) Mean-of-$K$ counterpart to \cref{fig:o3_bo_latents}: subspace dimension sweep across budgets.}
\label{fig:o3_appendix_complementary}
\end{figure}

\subsubsection*{O3 Ablate Budget Allocation}
\begin{figure}[H]
\vspace{-0.3cm}
\centering
\begin{subfigure}{0.49\linewidth}
  \centering
  \caption{}
  \includegraphics[width=\linewidth]{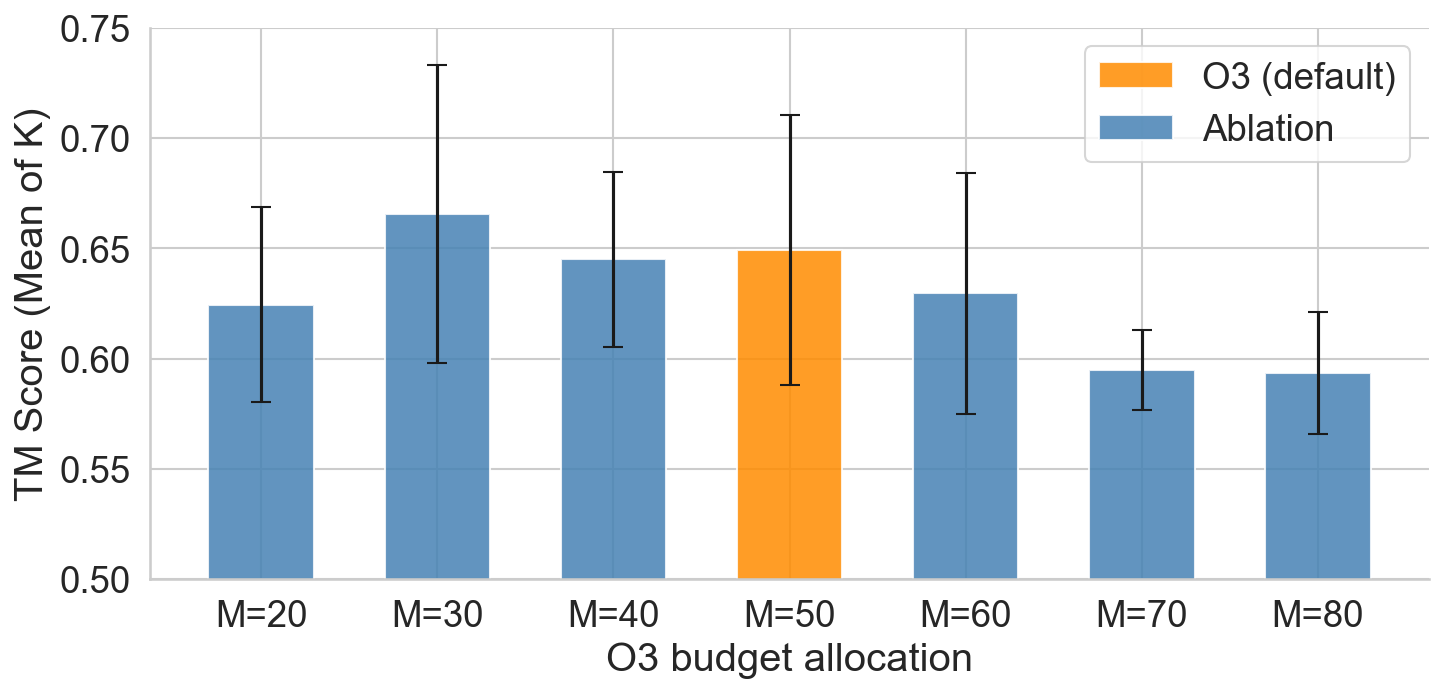}
  \label{fig:o3_ablate_M_mean}
\end{subfigure}
\hfill
\begin{subfigure}{0.49\linewidth}
  \centering
  \caption{}
  \includegraphics[width=\linewidth]{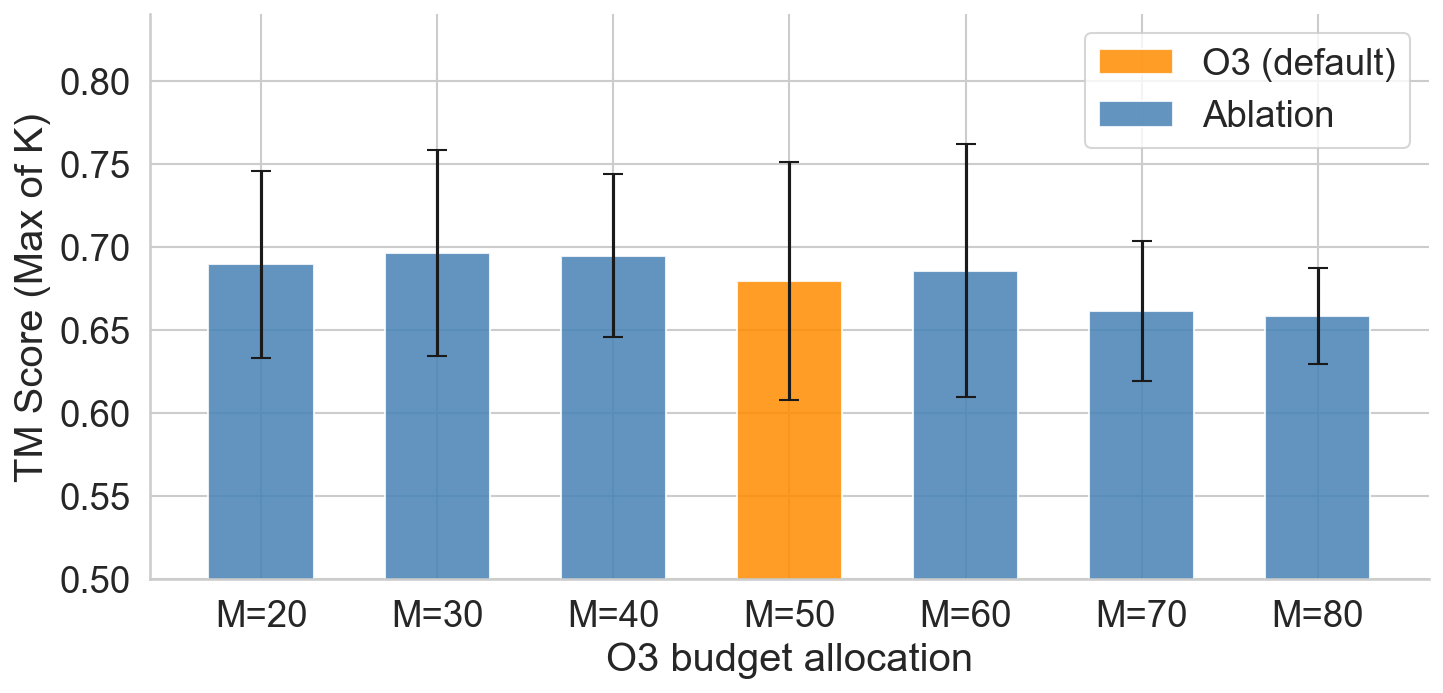}
  \label{fig:o3_ablate_M_max}
\end{subfigure}
\vspace{-0.3cm}
\caption{\textbf{O3 Ablation: Budget allocation M.} 
(a) Mean-of-$K$ TM Score reported for a fixed budget $N=100$, varying the initial number of samples used to create the $\mathcal{U}$ subspace as per \cref{sec:method:surrogate}, resulting in varying remaining BO budget, $N-M$ acquisition rounds. 
(b) Max-of-$K$ counterpart. 
In all cases we keep the subspace-dimension constant $d=7$, $K=10$, and draw $j=10$ uniform random $u$ samples as initial GP training data. Bars are the mean over 3 random seeds; error bars are $\pm 1$ std. Higher is better. Experiment uses a previous set of hyperparameters.}
\label{fig:o3_M_ablation}
\end{figure}

\subsubsection*{O3 Effect of GP Covariance function}
\begin{figure}[!htbp]
\vspace{-0.3cm}
\centering
\begin{subfigure}{0.49\linewidth}
  \centering
  \caption{}
  \includegraphics[width=\linewidth]{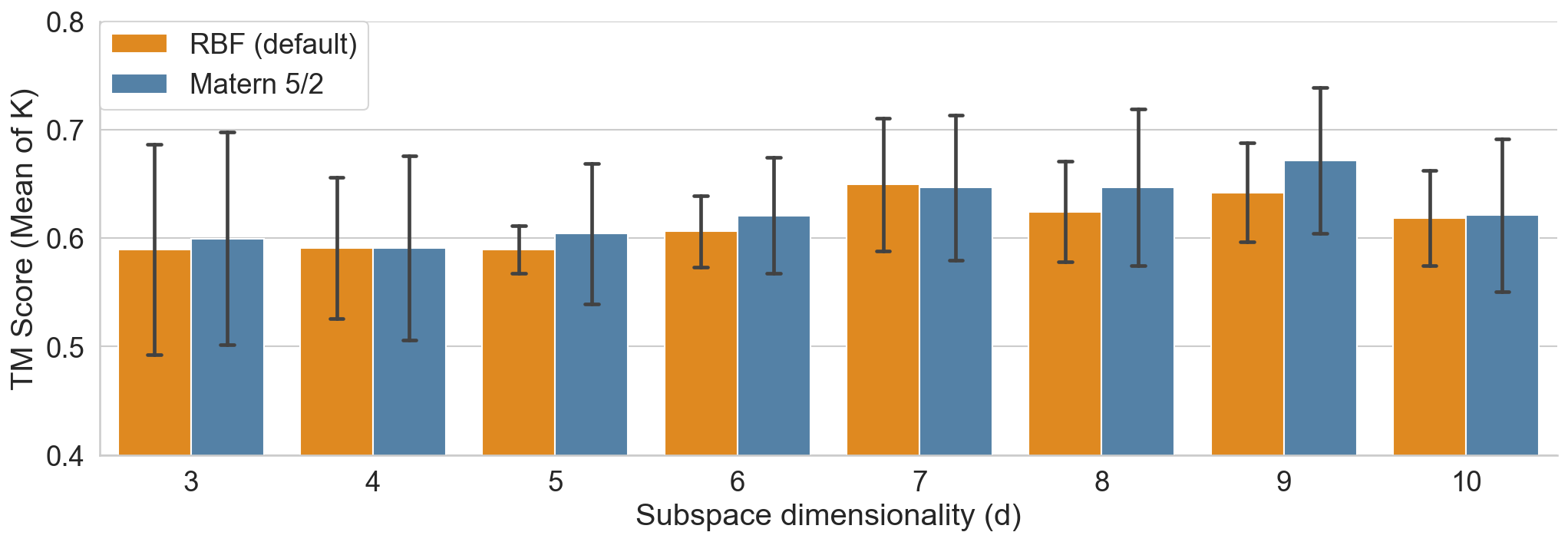}
  \label{fig:o3_ablate_kernel_mean}
\end{subfigure}
\hfill
\begin{subfigure}{0.49\linewidth}
  \centering
  \caption{}
  \includegraphics[width=\linewidth]{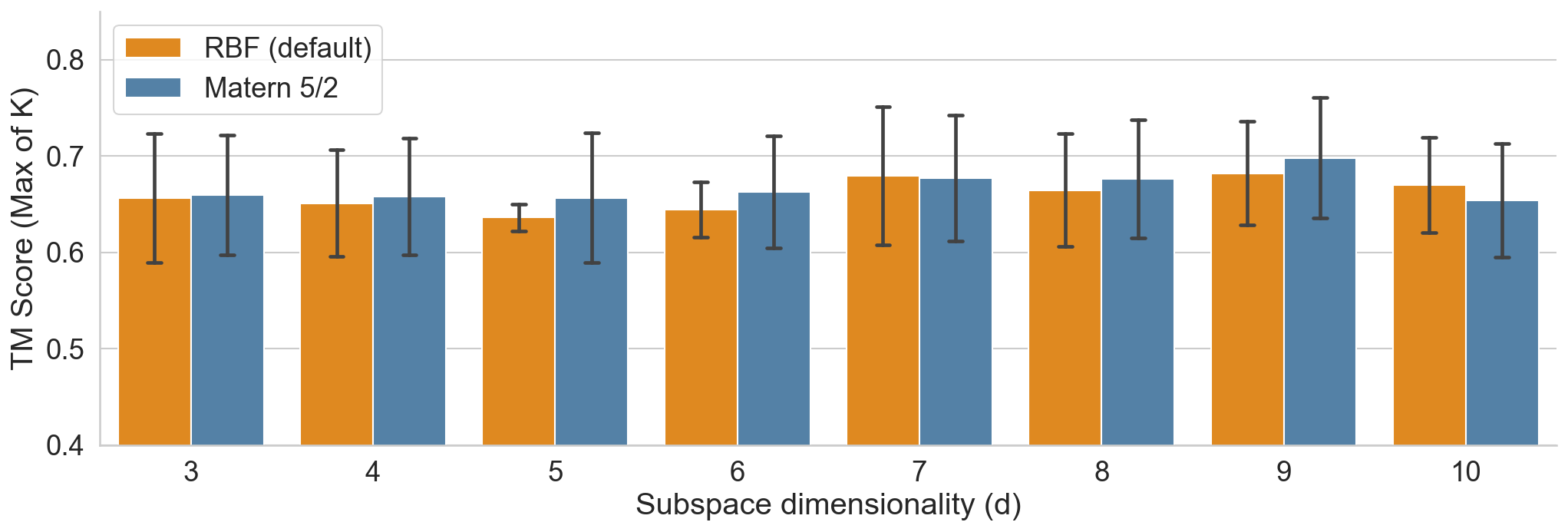}
  \label{fig:o3_ablate_kernel_max}
\end{subfigure}
\vspace{-0.3cm}
\caption{\textbf{O3 Ablation: GP covariance function.} 
(a) Mean-of-$K$ TM Score across different subspace dimensionality for an RBF kernel vs Matern 5/2 kernel as Gaussian process covariance function. 
(b) Max-of-$K$ counterpart to (a). 
In all cases we keep the budget $N=100$, batch size $K=10$, with $j=10$ uniform random $u$ samples, as per \cref{sec:method}. 
Bars are the mean over 5 random seeds; error bars are $\pm 1$ std. Higher is better. Experiment uses a previous set of hyperparameters.}
\label{fig:o3_kernel_ablation}
\end{figure}

\subsubsection*{O3 Effects of Seed selection strategy}
\begin{figure}[H]
\centering
\begin{subfigure}{0.49\linewidth}
  \centering
  \caption{}
  \includegraphics[width=\linewidth]{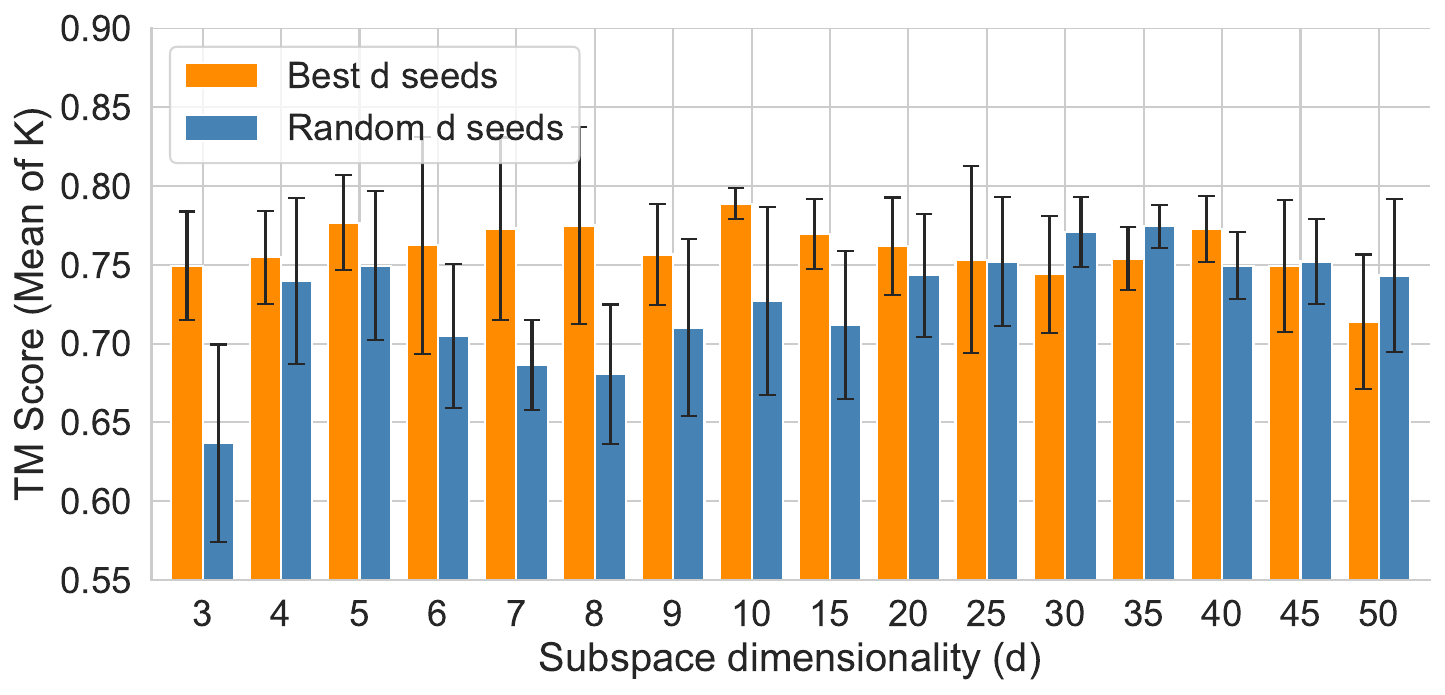}
  \label{fig:o3_ablate_seed_mean}
\end{subfigure}
\hfill
\begin{subfigure}{0.49\linewidth}
  \centering
  \caption{}
  \includegraphics[width=\linewidth]{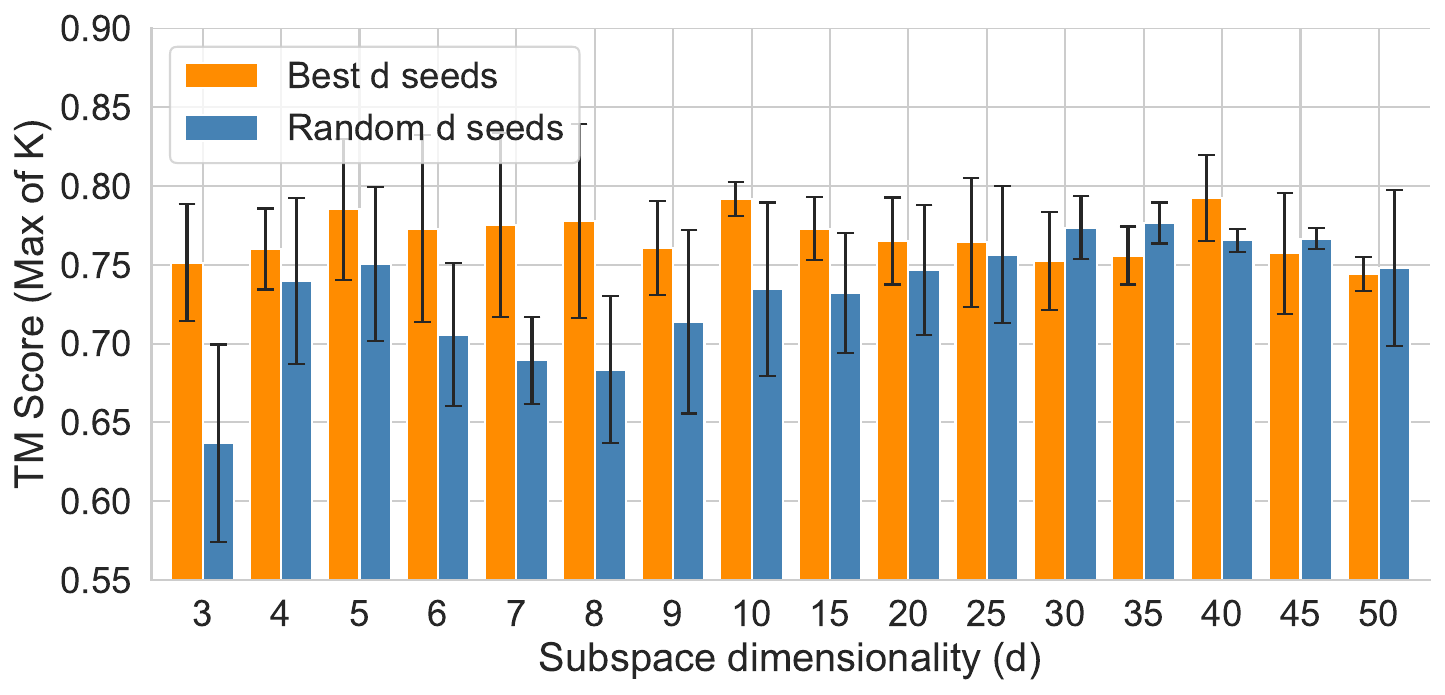}
  \label{fig:o3_ablate_seed_max}
\end{subfigure}
\caption{ \textbf{O3 Ablation: Choice of seed selection strategy.} 
(a) Mean-of-$K$ TM Score on 1CLL for different approaches to construct the $\mathcal{U}$ subspace: either the ``best $d$ seeds'' or ``random $d$ seeds'' are used, across a sweep of subspace dimensionalities.
Using random seeds means M = 0, so number of BO rounds is $N - (d+2)$, whereas best $d$ seeds has fewer $N - M - 2$. 
(b) Max-of-$K$ counterpart to (a).
In all cases we keep the budget $N=100$ and batch size $K=10$. 
Bars are the mean over 3 random seeds; error bars are $\pm 1$ std. Higher is better.
}
\label{fig:o3_seed_ablation}
\end{figure}

\subsection{Additional FK-Steering Results on PDB: 1CLL}\label{app:1cll:FK}
\subsubsection*{Complementary FK-Steering Metrics}
\begin{figure}[H]
    \centering
    \includegraphics[width=0.65\linewidth]{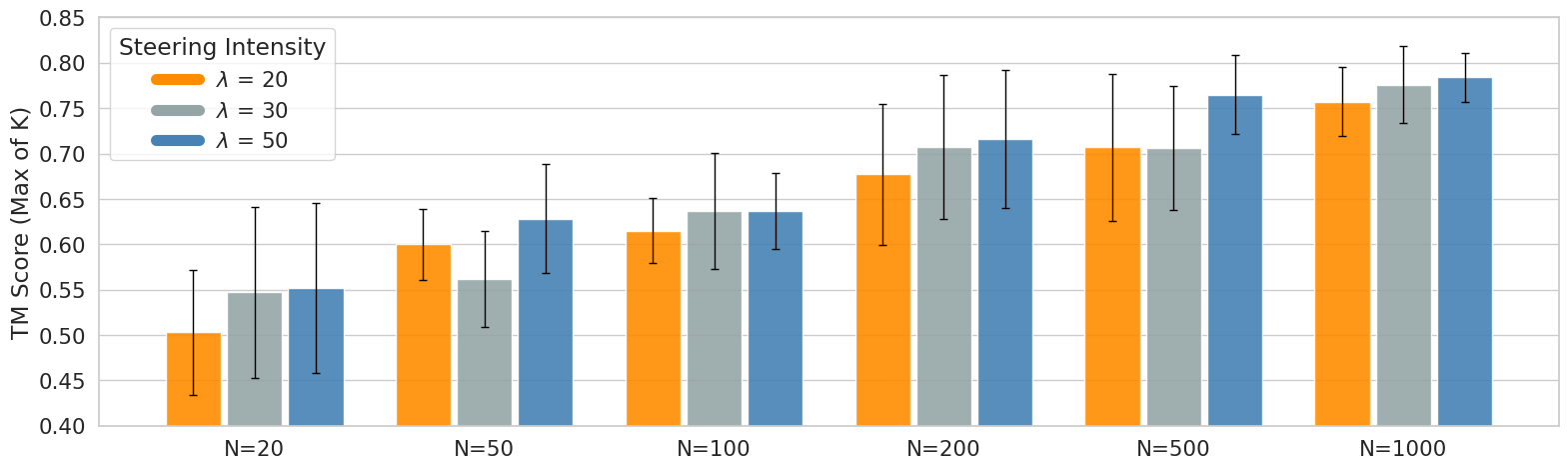}
    \caption{\textbf{Effect of FK-steering $\lambda$ hyperparameter across different budgets $N$}.  Max-of-K TM-score achieved by FK-steering on 1CLL. Higher $\lambda$ increases the signal from the rewards and leads to better performance across budgets. Bars are means over $10$ random seeds; error bars are $\pm 1$ std. }
    \label{fig:fk_lambda_max_k}
\end{figure}
\subsubsection*{FK-Steering $K$ vs $N/K$ tradeoff}
\begin{figure}[!htbp]
\centering
\begin{subfigure}{0.49\linewidth}
  \centering
  \caption{}
  \includegraphics[width=\linewidth]{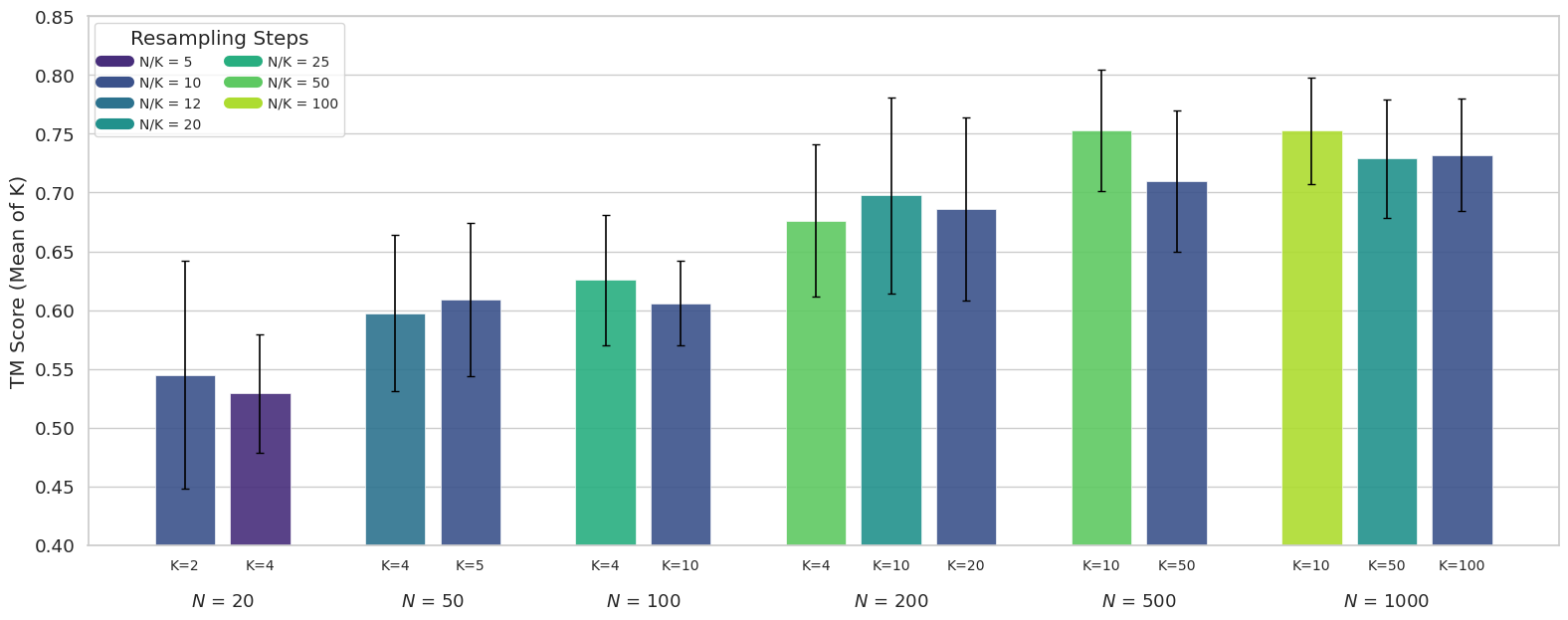}
  \label{fig:fk_NK_ablation_mean}
\end{subfigure}
\hfill
\begin{subfigure}{0.49\linewidth}
  \centering
  \caption{}
  \includegraphics[width=\linewidth]{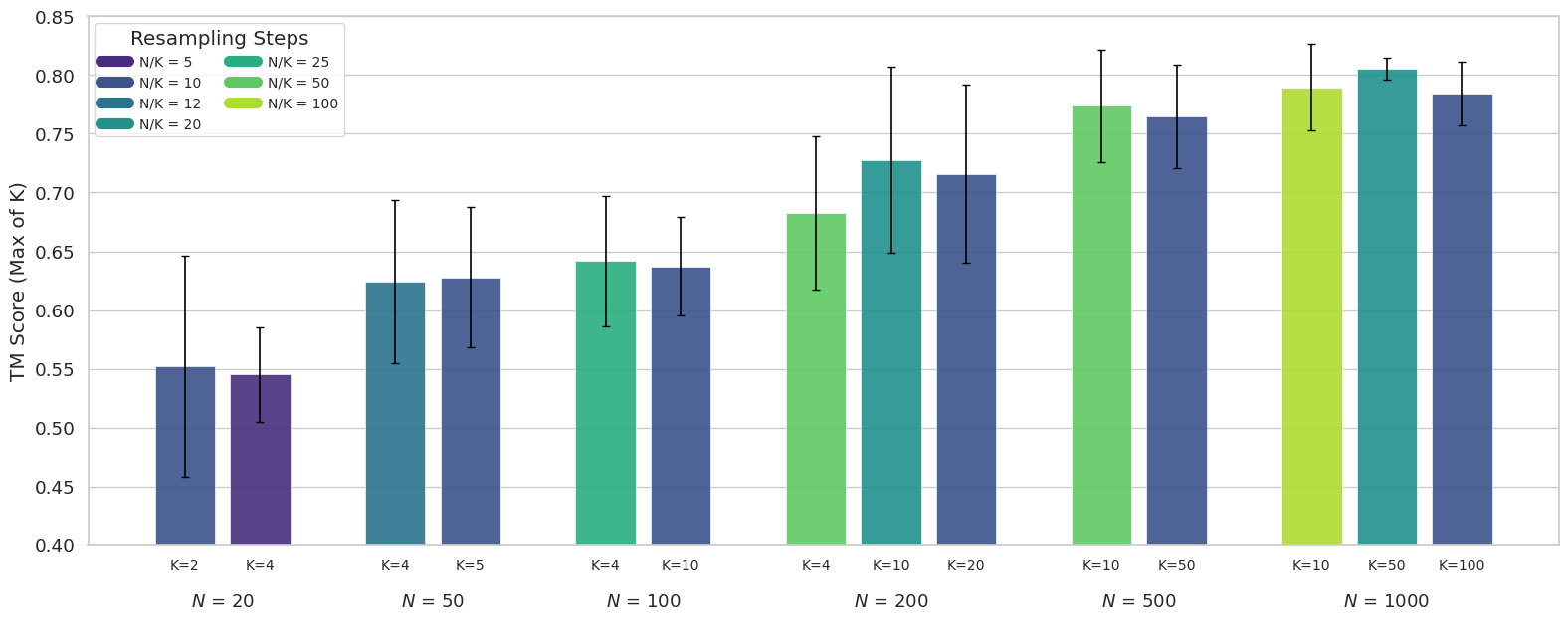}
  \label{fig:fk_NK_ablation_max}
\end{subfigure}
\caption{ \textbf{FK-steering $K$ vs $N/K$ tradeoff for a set budget.}
(a) Mean-of-$K$ TM Score for different approaches to allocate $K$ particles and $N/K$ resampling steps per set budget of $N$ oracle calls.
(b) Max-of-$K$ counterpart to (a). For a given budget $N$, increasing the number of resampling steps $N/K$ at the cost of particle population $K$ generally improves performance in the higher budget groups.
In all cases $\lambda=50$.
Bars are the mean over 10 random seeds; error bars are $\pm 1$ std. Higher is better.
}
\label{fig:fk_NK_ablation}
\end{figure}

\subsubsection*{FK-Steering reward trajectories}
\begin{figure}[H]
\centering
\begin{subfigure}{0.49\linewidth}
  \centering
  \caption{}
  \includegraphics[width=\linewidth]{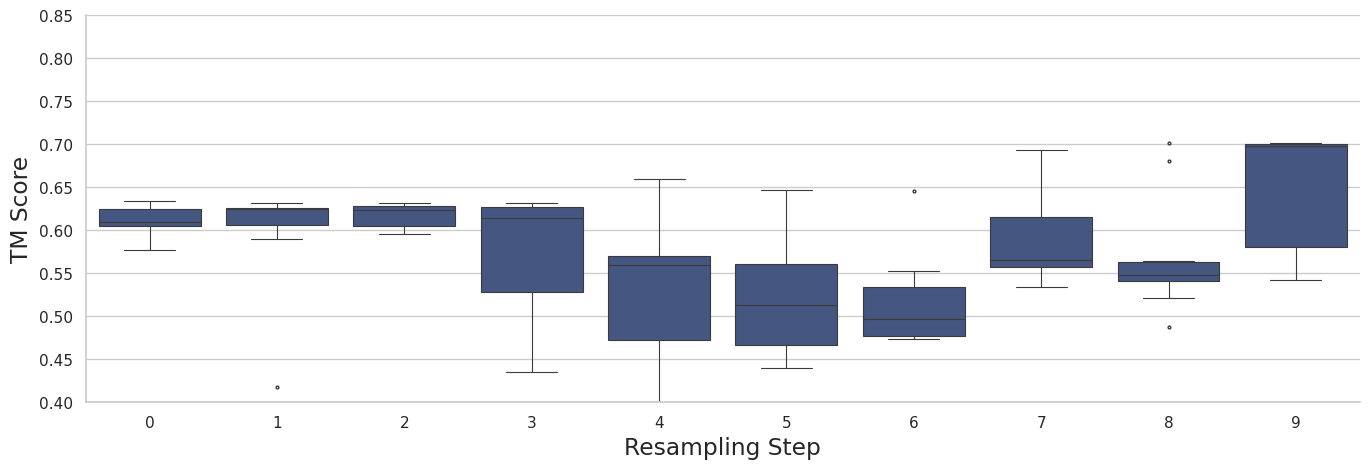}
  \label{fig:fk_NK_10}
\end{subfigure}
\hfill
\begin{subfigure}{0.49\linewidth}
  \centering
  \caption{}
  \includegraphics[width=\linewidth]{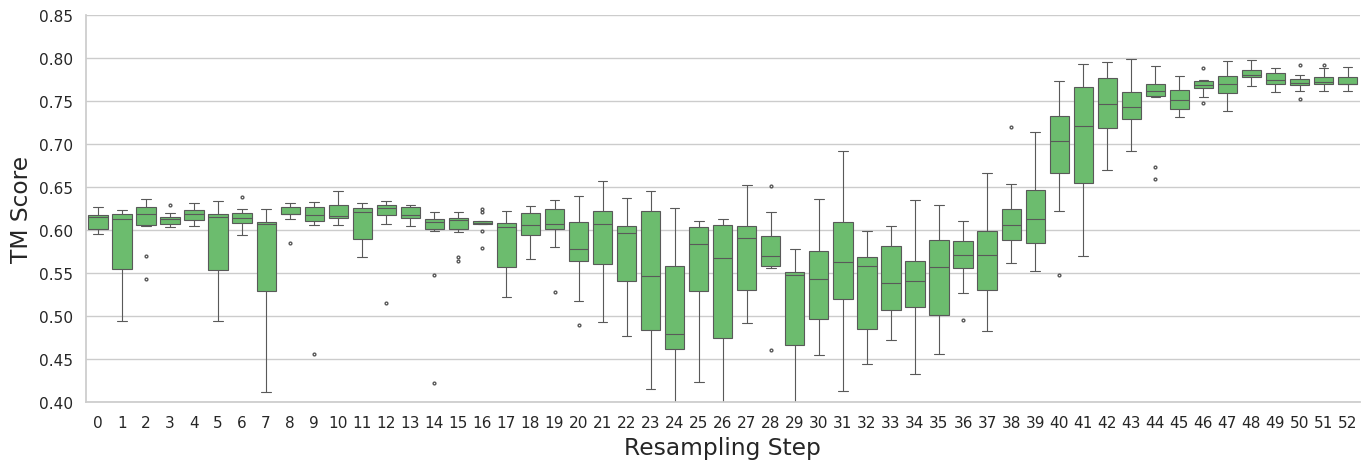}
  \label{fig:fk_NK_50}
\end{subfigure}
\caption{\textbf{FK-steering reward trajectory for $K=10$ particles with different number of resampling steps $N/K$}. (a) $N/K=10$ resampling steps with a budget of $N=100$ oracle calls. 
(b)  $N/K=50$ resampling steps with a budget of $N=500$ oracle calls. 
In both cases $\lambda=50$. Colors preserve $N/K$ colorscheme from \cref{fig:fk_NK_ablation}. Resampling more frequently at higher $N/K$ propagates the reward signal better and reduces reward variance in the final resampling steps.}
\label{fig:fk_NK_illustration}
\end{figure}

\subsection{Additional DPO Results on PDB: 1CLL}\label{app:1cll:dpo}
\begin{figure*}[h]
\centering
\includegraphics[width=0.5\linewidth]{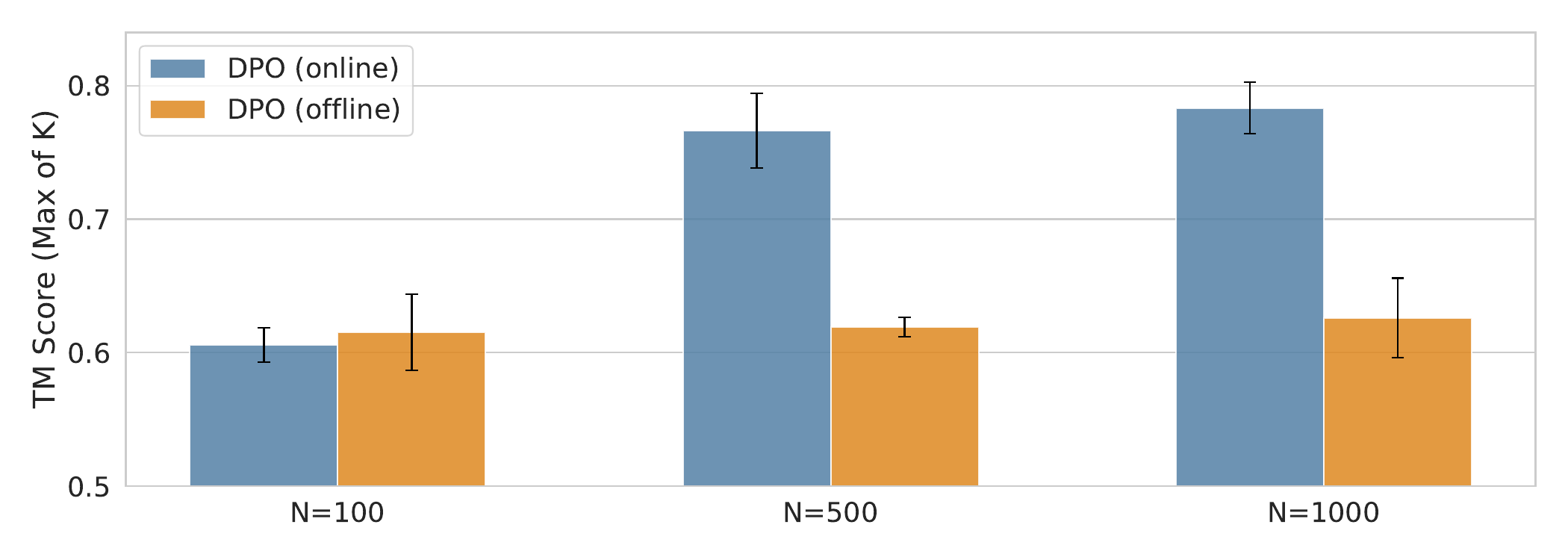}
\caption{
\textbf{Online vs.\ offline DPO across oracle budgets.} Max-of-$K$ TM-score achieved by online and offline DPO on 1CLL, for oracle budgets $N \in \{100, 500, 1000\}$. Bar are means over 5 random seeds; error bars show $\pm 1$ std. Online DPO improves consistently with $N$, while offline DPO remains flat across budgets.
}
\label{fig:dpo-online-offline-max}
\end{figure*}

\newpage
\subsection{Compute resources used in PDB: 1CLL Experiments}
\begin{table*}[h]
\centering
\caption{
\textbf{Compute resources used by the Boltz-2 guidance PDB: 1CLL experiment reported in this paper.} `Evals/run' counts oracle (TM-score) evaluations, each of which corresponds to one full generation through Boltz-2. The FK-steering rows aggregate over the $3$ values of $\lambda$ swept per budget.
}
\resizebox{\textwidth}{!}{%
\begin{tabular}{@{}lllrrrl@{}}
\toprule
Experiment & Model / pipeline & Hardware & Evals/run & \# runs & Per-run wall-clock & Total wall-clock \\
\midrule
O3~\citep{willis2025defining} & \textsc{Boltz-2} (PF-ODE) & H100 & 6 budgets up to $N{=}1000$ & 5 per budget & $\sim 0.1$ h & $\sim 3$ h \\
      &     &    & $N{=}20$ & 5 & $2.7$ m & $13$ m \\
      &     &    & $N{=}50$ & 5 & $2.8$ m & $14$ m \\
      &     &    & $N{=}100$ & 5 & $3.1$ m & $16$ m \\
      &     &    & $N{=}200$ & 5 & $2.9$ m & $15$ m \\
      &     &    & $N{=}500$ & 5 & $3.4$ m & $17$ m \\
      &     &    & $N{=}1000$ & 5 & $18$ m & $1$ h $31$ m \\
Best K-of-N & \textsc{Boltz-2} & H100 & 6 budgets up to $N{=}1000$ & 5 per budget & $\sim 0.1$ h & $\sim 3$ h \\
      &     &    & $N{=}20$ & 5 & $2.7$ m & $13$ m \\
      &     &    & $N{=}50$ & 5 & $2.8$ m & $14$ m \\
      &     &    & $N{=}100$ & 5 & $3.1$ m & $16$ m \\
      &     &    & $N{=}200$ & 5 & $2.9$ m & $15$ m \\
      &     &    & $N{=}500$ & 5 & $3.4$ m & $17$ m \\
      &     &    & $N{=}1000$ & 5 & $18$ m & $1$ h $31$ m \\
FK-steering~\citep{singhal2025general} & \textsc{Boltz-2} & H100 & 6 budgets, $3$ $\lambda$ values each &  &  &  \\
      &     &    & $N{=}20$, $3$ $\lambda$ vals & 10 & $3.3$ m & $1$ h $39$ m \\
      &     &    & $N{=}50$, $3$ $\lambda$ vals & 10 & $3.8$ m & $1$ h $54$ m \\
      &     &    & $N{=}100$, $3$ $\lambda$ vals & 10 & $4.1$ m & $2$ h $3$ m \\
      &     &    & $N{=}200$, $3$ $\lambda$ vals & 10 & $5$ m & $2$ h $30$ m \\
      &     &    & $N{=}500$, $3$ $\lambda$ vals & 10 & $6.9$ m & $3$ h $27$ m \\
      &     &    & $N{=}1000$, $3$ $\lambda$ vals & 10 & $11$ m & $5$ h $30$ m \\
Online DPO~\citep{rafailov2023direct,wallace2023diffusion} & \textsc{Boltz-2} (fine-tuned) & H100 & 6 budgets up to $N{=}1000$ &  &  &  \\
      &     &    & $N{=}20$ & 4 & $4$ m & $16$ m \\
      &     &    & $N{=}50$ & 4 & $6$ m & $24$ m \\
      &     &    & $N{=}100$ & 5 & $11$ m & $55$ m \\
      &     &    & $N{=}200$ & 3 & $17$ m & $51$ m \\
      &     &    & $N{=}500$ & 5 & $44$ m & $3$ h $40$ m \\
      &     &    & $N{=}1000$ & 4 & $1$ h $31$ m & $6$ h $04$ m \\
Offline DPO~\citep{rafailov2023direct,wallace2023diffusion} & \textsc{Boltz-2} (fine-tuned) & H100 & 6 budgets up to $N{=}1000$ &  &  &  \\
      &     &    & $N{=}20$ & 4 & $4$ m & $16$ m \\
      &     &    & $N{=}50$ & 5 & $7$ m & $35$ m \\
      &     &    & $N{=}100$ & 4 & $18$ m & $1$ h $12$ m \\
      &     &    & $N{=}200$ & 5 & $56$ m & $4$ h $40$ m \\
      &     &    & $N{=}500$ & 4 & $5$ h $40$ m & $22$ h $40$ m \\
      &     &    & $N{=}1000$ & 3 & $21$ h $57$ m & $65$ h $51$ m \\
\bottomrule
\end{tabular}
}
\label{tab:compute}
\end{table*}

\section{Experimental Setup and Results on PDB: 9EEH}\label{app:9eeh}
\subsection{MolProbity Oracle}\label{app:9eeh:MP}
TM-score (\cref{sec:setup}) requires a ground-truth structure, which is usually unavailable in the design settings this paper targets. We therefore also evaluate against a reference-free oracle that scores the physical plausibility of a generated structure, built on MolProbity \citep{chen2010molprobity, williams2018molprobity}. The oracle runs six checks. Backbone dihedral quality comes from a Ramachandran analysis and C$\beta$ deviation count via \texttt{mmtbx}. Steric quality is the clashscore, the number of serious all-atom overlaps (\texttt{probe} ``bad overlap`` contacts with a gap below $-0.4$~\AA) per 1000 atoms, computed after adding and optimising all hydrogens with \texttt{reduce}. Covalent geometry is scored as RMS $Z$-scores of backbone bond lengths and bond angles against the Engh-Huber reference values, and peptide planarity contributes counts of cis-nonPro and twisted $\omega$ angles. These terms are combined into a single log-weighted penalty following the MolProbity score formulation, which we flip and normalise to a validity score in $[0, 1]$, with higher values indicating fewer physical violations. 

\subsection{PDB: 9EEH Results}\label{app:9eeh:results}
We repeat the experiment of \cref{sec:results:budgets} on E. coli aspartate transcarbamoylase (PDB: 9EEH)~\cite{miller2026}, scored with the reference-free MolProbity oracle described above. There are three key differences between this setup and the one with the TM-score oracle on the calmodulin protein (PDB: 1CLL). First, 9EEH was deposited to the PDB after the training cutoff of Boltz-2, thus it is likely an out-of-distribution task for the base model, whereas 1CLL is not. Second, 9EEH is a 7,232-atom complex, which is much larger than 1CLL with 1,184 atoms. Third, the MolProbity score measures physical plausibility rather than similarity to a reference structure, and its effective range is a lot narrower, with most structures typically scoring between 0.5 and 0.7 under our MolProbity implementation.

\begin{figure*}[h]
\centering
\includegraphics[width=0.75\linewidth]{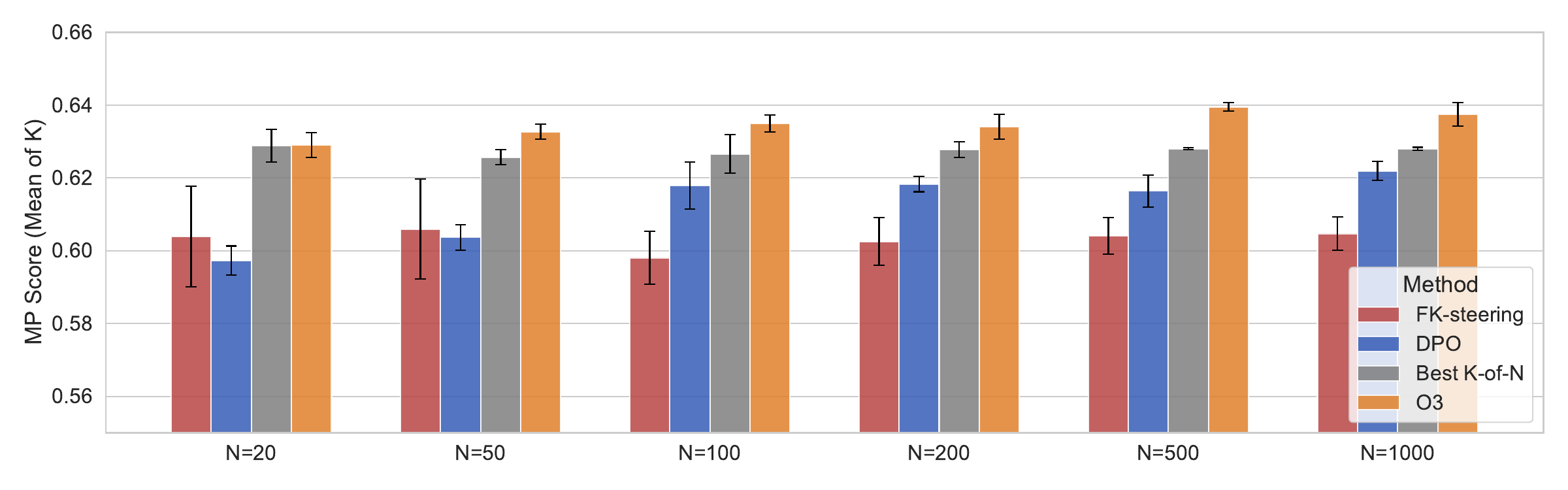}
\caption{\textbf{Mean-of-$K$ MolProbity score across oracle budgets.} Comparison of FK-steering, DPO (in the online setting), Best K-of-N, and O3 on 9EEH across the six $(N, K)$ configurations of \cref{sec:setup}. Bars are the mean over $3$ random seeds; error bars are $\pm 1$ std. Higher is better.}
\label{fig:mean_top_k_across_n_9eeh}
\vspace{-0.2cm}
\end{figure*}

\begin{figure*}[h]
\vspace{-0.5cm}
\centering
\includegraphics[width=0.75\linewidth]{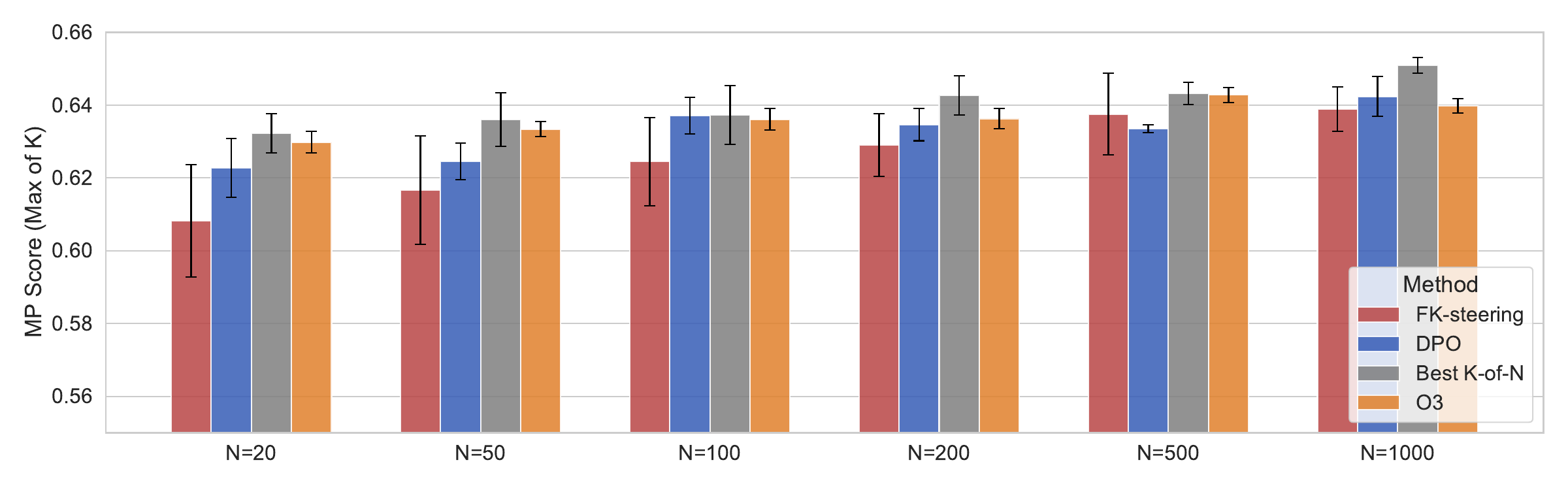}
\caption{\textbf{Max-of-$K$ MolProbity score across oracle budgets.} Comparison of FK-steering, DPO, Best K-of-N, and O3 on 9EEH across the six $(N, K)$ configurations of \cref{sec:setup}, reporting the maximum MolProbity score across the returned batch of $K$ candidate structures. Bars are the mean over $3$ random seeds; error bars are $\pm 1$ std. Higher is better.}
\label{fig:top_k_max_appendix_9eeh}
\vspace{-0.2cm}
\end{figure*}

\cref{fig:mean_top_k_across_n_9eeh} reports the mean-of-$K$ MolProbity score. O3 scores highest at all but the smallest budget, where it matches Best K-of-N. Best K-of-N has a relatively constant performance across budgets, as expected under fixed $K/N$ ratio. DPO improves with N, although it performs significantly worse than Best K-of-N at low budgets. FK-steering performs poorly throughout and does not improve with budget, in contrast to 1CLL where it does better than Best K-of-N for budgets above $N = 100$. In \cref{fig:top_k_max_appendix_9eeh} under max-of-$K$, Best K-of-N is strongest at large budgets and both it and FK-steering improve with $N$, while O3 is unchanged.

Under max-of-$K$, O3 performs worse than Best K-of-N, and worse than all methods at the highest budget. We hypothesise that this is because it is the only method that requires sampling Boltz-2 with the probability-flow ODE while the other three methods use the stochastic Boltz-2 sampler which improves the output diversity. Under mean-of-$K$ this yields worse performance overall because the lower-scoring samples drag down the average, however it improves the max-of-$K$ scores, where only the best sample matters and increasing $N$ gives better scores even without any guidance signal (i.e. Best K-of-N baseline).

FK-steering is the only method that scores intermediate denoised predictions, and it is the weakest method under the MolProbity oracle. TM-score varies smoothly with structural accuracy: the same global fold already scores above 0.5, and refinement moves the score toward 1, so intermediate predictions receive informative scores. MolProbity instead measures local geometry, in which residual noise dominates until the final denoising steps. Since the quality of the denoised predictions is much worse at the intermediate steps than at the final ones, this leaves noise-sensitive intermediate MolProbity scores nearly uninformative. Moreover, since FK-steering relies on a positive noise scale for particle diversity, in Boltz-2 resampling is concentrated in the first three quarters of the denoising trajectory, which exacerbates the issues with signal quality from the MolProbity oracle at intermediate steps.

Finally, the differences between methods on this target are small, and the error bars over three seeds overlap at several budgets. Three of the four methods behave much as they do on 1CLL — O3 scores highest, DPO improves with $N$, and Best K-of-N is relatively constant — with FK-steering the exception, for the reason above. We therefore read these orderings as consistent trends across budgets rather than as individually meaningful differences, and note that guidance is less effective on this target than on 1CLL with TM-score.

\subsection{Additional O3 Details and Results on PDB: 9EEH}\label{app:9eeh:results:o3}

\subsubsection*{O3 Experimental Details}
\begin{table}[H]
  \centering
  \caption{%
  Experimental configurations for O3 protein structure optimisation
      (Boltz-2 generator, MolProbity oracle on 9EEH).
      $N$: total Oracle budget;
      $K$: top-$K$ batch selection;
      $M$: initial samples filtered to create subspace $\mathcal{U}$;
      $d$: number of seed latents used to create the $d-1$ dimensional $\mathcal{U}$;
      $d+2$: initial points used to fit the Gaussian process in $\mathcal{U}$;
      $n_{\text{rounds}}$: number of BO acquisition rounds is set to $N-M-2$ for filtered seed-latents (and $N-(d+2)$ for random seed latents); 
      Wall-clock time of experiments run using a H100 GPU with 80Gb RAM. 
  }
  \label{tab:boltz_tm_configs}
  \begin{tabular}{lrr| rrrc}
  \toprule
  Task & $N$ & $K$ & $M$ & d & $n_{\text{rounds}}$  & wall-clock time (mins) \\
  \midrule
  \texttt{n20\_k2}     &    20 & 2 & 10 & 5 & 8   &   9.1 ($\pm$0.1)\\
  \texttt{n50\_k5}     &    50 & 5 & 25 & 5 & 23  &  19.2 ($\pm$0.3)\\
  \texttt{n100\_k10}   &   100 & 10 & 50 & 20 & 48 &  40.0 ($\pm$ 0.7) \\
  \texttt{n200\_k20}   &   200 & 20 & 100 & 20 & 98  &  75.9 ($\pm$2.0)\\
  \texttt{n500\_k50}   &   500 & 50 & 250 & 30 & 248 &  224.2 ($\pm$9.0)\\
  \texttt{n1000\_k100} &  1000 & 100 & 400 & 50 & 598  &  368.0 ($\pm$7.4)\\
  \bottomrule
  \end{tabular}
\end{table}

\subsubsection*{O3 Effects of Seed selection strategy}
\begin{figure}[H]
\vspace{-0.3cm}
\centering
\begin{subfigure}{0.49\linewidth}
  \centering
  \caption{}
  \includegraphics[width=\linewidth]{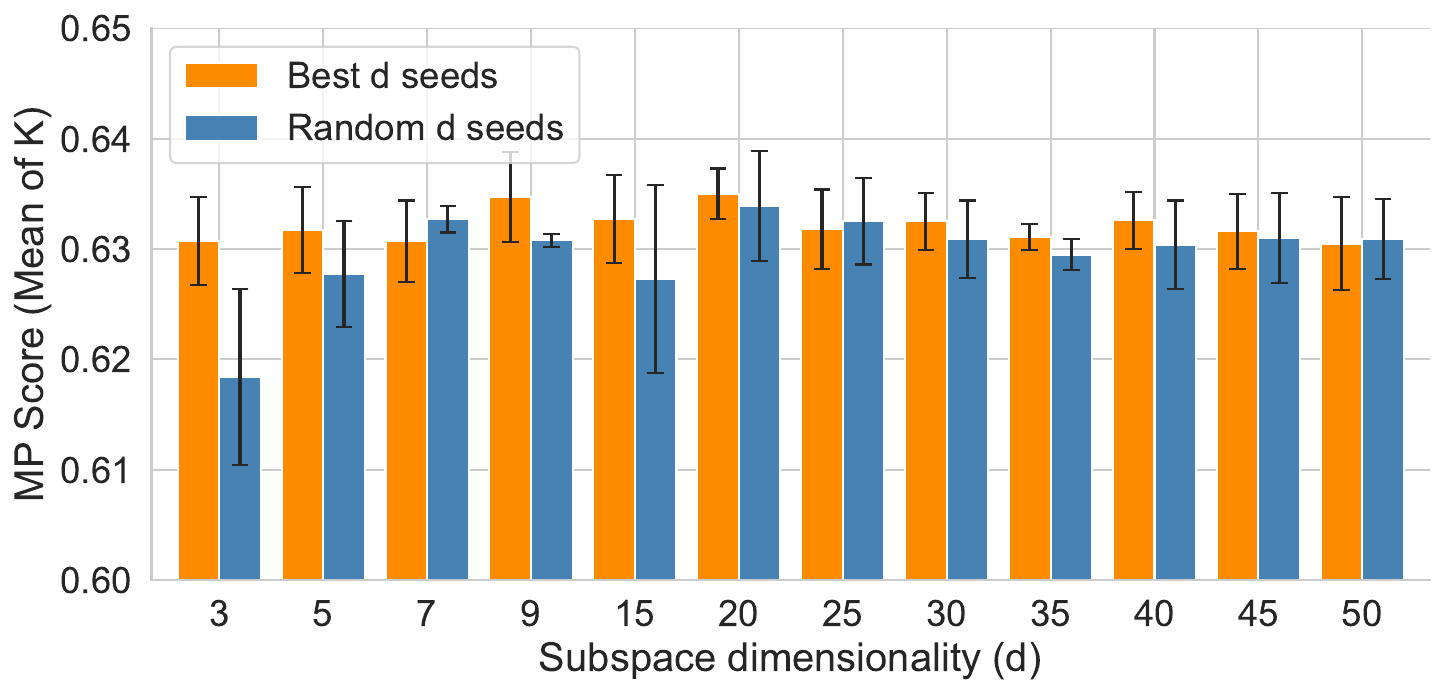}
  \label{fig:o3_ablate_seed_mean}
\end{subfigure}
\hfill
\begin{subfigure}{0.49\linewidth}
  \centering
  \caption{}
  \includegraphics[width=\linewidth]{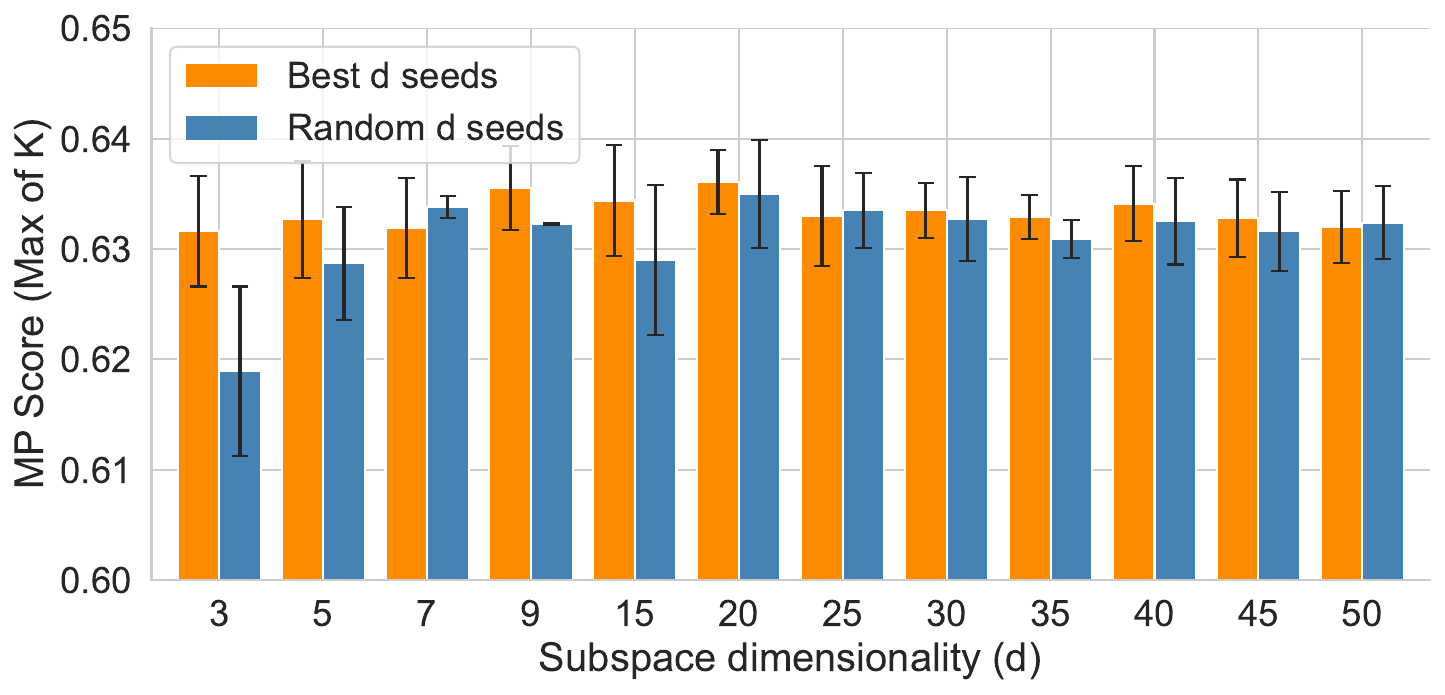}
  \label{fig:o3_ablate_seed_max}
\end{subfigure}
\vspace{-0.3cm}
\caption{ \textbf{O3 Ablation: Choice of seed selection strategy.} 
(a) Mean-of-$K$ MolProbity Score on 9EEH for different approaches to construct the $\mathcal{U}$ subspace: either the ``best $d$ seeds'' or ``random $d$ seeds'' are used, across a sweep of subspace dimensionalities. 
Using random seeds means M = 0, so the number of BO rounds is $N - (d+2)$, whereas best $d$ seeds has fewer $N - M - 2$. 
(b) Max-of-$K$ counterpart to (a).
In all cases, we keep the budget $N=100$ and batch size $K=10$. 
Bars are the mean over 3 random seeds; error bars are $\pm 1$ std. Higher is better.
}
\label{fig:o3_seed_ablation}
\end{figure}


\end{document}